\documentclass[journal]{IEEEtran}

\ifCLASSINFOpdf
\else
   \usepackage[dvips]{graphicx}
\fi
\usepackage{url}

\usepackage{multirow,diagbox,algorithm}
\usepackage{amsmath,amssymb,amsfonts}
\usepackage{algorithmic}
\usepackage{graphicx}
\usepackage{textcomp}
\newcommand{\reffig}[1]{Fig. \ref{#1}}
\newcommand{\reftable}[1]{Table \ref{#1}}
\newcommand{\refeq}[1]{Eq. \ref{#1}}

\begin{document}
	
	\title{Unsupervised Many-to-Many Image-to-Image Translation Across Multiple Domains}
	
	\author{Ye Lin, Keren Fu, Shenggui Ling and Peng Cheng*
	\thanks{*Corresponding author. }
	\thanks{This work was supported by the Joint Funds of the National Natural Science Foundation of China under Grant U1833128, and was partly supported by the National Science Foundation of China, under No. 61703077, the Fundamental Research Funds for the Central Universities No. YJ201755, and the Sichuan Science and Technology Major Projects (2018GZDZX0029).}
	\thanks{Y. Lin (e-mail: linlanye@sina.cn) and S. Ling (710019969@qq.com) are with the National Key Laboratory of Fundamental Science on Synthetic Vision, Sichuan University, Chengdu, Sichuan 610065, China.}
	\thanks{K. Fu (fkrsuper@scu.edu.cn) is with the College of Computer Science, Sichuan University, Chengdu, Sichuan 610065, China.}
	\thanks{P. Cheng (chengpeng\_scu@163.com) is with the School of Aeronautics and Astronautics, Sichuan University, Chengdu, Sichuan 610065, China.}
	}
	
	\markboth{Journal of \LaTeX\ Class Files, Vol. 14, No. 8, August 2015}
	{Shell \MakeLowercase{\textit{et al.}}: Bare Demo of IEEEtran.cls for IEEE Journals}
	\maketitle
	
	\begin{abstract}
		Unsupervised multi-domain image-to-image translation aims to synthesis images among multiple domains without labeled data, which is more general and complicated than one-to-one image mapping. However, existing methods mainly focus on reducing the large costs of modeling and do not pay enough attention to the quality of generated images. In some target domains, their translation results may not be expected or even it has model collapse. To improve the image quality, we propose an effective many-to-many mapping framework for unsupervised multi-domain image-to-image translation. There are two key aspects in our method. The first is a proposed many-to-many architecture with only one domain-shared encoder and several domain-specialized decoders to effectively and simultaneously translate images across multiple domains. The second is two proposed constraints extended from one-to-one mappings to further help improve the generation. All the evaluations demonstrate our framework is superior to existing methods and provides an effective solution for multi-domain image-to-image translation.
	\end{abstract}
	

	\IEEEpeerreviewmaketitle

\section{INTRODUCTION}
In image processing, image-to-image translation is regarded as an one-to-one mapping issue between two different image domains, which enables an input image to obtain features of another desired domain \cite{isola2017image}. It is widely applied in computer vision industry such as, image segmentation \cite{papandreou2015weakly}, style transfer \cite{johnson2016perceptual}, image colorization \cite{zhang2016colorful}, face synthesis \cite{Abboud2003Expressive} \cite{kazemi2018facial}, image inpainting \cite{pathak2016context} \cite{Xiao2018Pattern} and super-resolution \cite{ledig2017photo}. Without needing paired images, the unsupervised image-to-image translation methods \cite{zhu2017unpaired} \cite{yi2017dualgan} \cite{liu2017unsupervised} \cite{kim2017learning} \cite{taigman2017unsupervised} are more applicable than supervised ones \cite{mirza2014conditional} \cite{isola2017image}, since data preparation only involves dividing the images into different domains. For translations among multiple domains denoted as $N$, namely multi-domain image-to-image translation, they are impractical due to the $O(N^2)$ times of mutual training effort to achieve $N(N-1)$ different mappings.

To simplify the training process for multi-domain image-to-image translation, a variety of schemes are adopted. One type of representatives \cite{anoosheh2018combogan} \cite{hui2018unsupervised} is to divide each mapping into two separate processes, in which they use $N$ domain-specialized encoders to separately process corresponding images from different domains and employ $N$ domain-specialized decoders to finish the generations in different target domains. Thus, it only needs to choose the correct encoder and decoder to combine to complete the final translation. Another scheme  \cite{choi2018stargan} \cite{tang2018dual} \cite{lin2019relgan} \cite{wang2019sdit} \cite{li2019asymmetric} is to utilize a single generator with an auxiliary domain label vector to control translations among all domains.  Although all these solutions can efficiently process multi-domain image translation, the quality of their translated images are not always good enough. There are still difficulties that cause the performance of their models degrade. For the methods with splitting the generation, they have enough capacity to handle more general translations across any number of domains, but they are difficult to balance the training process between different domain pairs (input domain to output domain), possibly resulting in poor image quality or even model collapse. For the methods with using an auxiliary vector, if the vector is not accurate enough to label different domains, the generated images may not conform to the target domain style. In addition, their single generator architecture makes them difficult to simultaneously handle the embeddings of different target domains, especially when there are large domain differences, possibly leading to unsatisfactory results.

\begin{figure}
	\begin{center}
		\includegraphics[width=0.9\hsize]{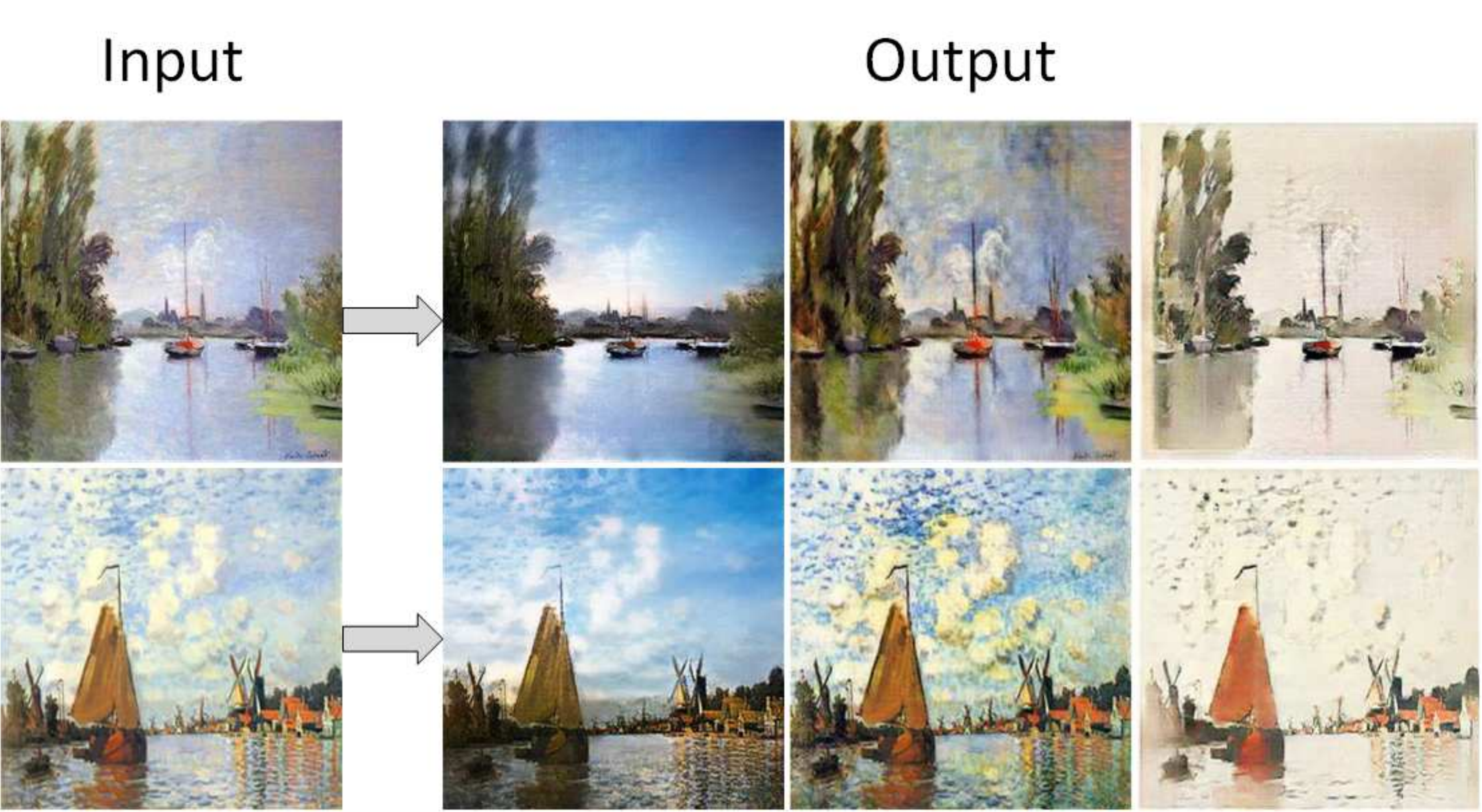}
	\end{center}
	\caption{An example of image translation using MDT which simultaneously generates images in all domains regardless of the input domain label.}
	\label{figs:example}
\end{figure}

To avoid the defects of the schemes mentioned above and improve the generations across all target domains, we present an unsupervised many-to-many image-to-image translation framework called multi-domain translator (MDT), based  on generative adversarial networks (GANs) \cite{goodfellow2014generative}. MDT has only one domain-shared encoder to reduce the interference of domain-specialized information from different source domains, and has multiple identical domain-specialized decoders to translate images across all target domains. We also propose two general constraints extended from one-to-one mappings \cite{zhu2017unpaired} \cite{yi2017dualgan} \cite{kim2017learning} to further help the multi-domain generation. The one named reconstruction constraint requires that, when the model generates images in all domains from a single input image, the generated domain-specialized images (except the result in the source domain) could be recovered to their original appearances in their source domain by feeding them to another cycled translation. The other enforces the input images to remain unchanged when they are processed by the decoders corresponding to their source domains, named identity consistency. With this architecture, MDT is easy to balance training and has sufficient capacity to process translations with large domain differences. An example of using MDT is shown in \reffig{figs:example}. 

Experiments are conducted on two types of image translation tasks with or without ground truth. Both qualitative and quantitative comparative evaluations demonstrate that MDT performs favorably against state-of-the-art methods \cite{anoosheh2018combogan} \cite{choi2018stargan} \cite{wang2019sdit} and offers an effective solution for unsupervised multi-domain image-to-image translation. In summary, our main contributions are listed as follows.
\begin{itemize}
	\item We propose an unsupervised many-to-many framework to better improve the image quality of multi-domain image-to-image translation.
	\item Two general constraints are proposed to further improve the translations across all domains.
	\item We analyze the advantages and disadvantages of existing modeling schemes to solve multi-domain image-to-image translation problems.				
	\item The potential and effectiveness of our framework is validated on different multi-domain image translation tasks through qualitative and quantitative evaluations.
\end{itemize}

The reminder of the paper is organized as follows. Section \ref{sec2} describes related work on image-to-image translation. Section \ref{sec3} introduces the proposed MDT in details. The implementation is presented in Section \ref{sec4}. Experiments and analyses are included in Section \ref{sec5}. Conclusions are drawn in Section \ref{sec6}. More applications of the proposed framework are presented in supplementary material.

\section{RELATED WORK}\label{sec2}
\subsection{Image-to-Image Translation in Two Domains}
GANs \cite{goodfellow2014generative} are widely used in solving image-to-image translation problems. The supervised approaches \cite{mirza2014conditional} \cite{isola2017image} based on GANs train their model with paired data to generate high-quality images in desired domain. However, collecting labeled data is difficulty, so unsupervised setting is favored by many researches \cite{kim2017learning} \cite{zhu2017unpaired} \cite{yi2017dualgan} which aim at achieving the same or even better results than the supervised ones. An effective practice is to utilize two generators to cooperate with each other to constraint and complete the entire image generation, which can be described as cycle consistency. The cycle consistency is an important constraint that requires the original image can be restored after successive two mappings by two different generators. Recent works, such as DiscoGAN \cite{kim2017learning}, CycleGAN \cite{zhu2017unpaired} and DualGAN \cite{yi2017dualgan}, use cycle consistency to improve the quality of generated images. Shen et.al \cite{shen2019one} further improve this constraint to obtain accurate one-to-one mapping. On the other hand, Liu and Tuzel \cite{liu2016coupled} propose CoGAN with tied weights on the first few layers for shared latent representation. Based on the idea of style transfer \cite{johnson2016perceptual}, Liu et al. use two pairs of encoders and decoders to embed input images into a same latent space and restore them to the target domain, called UNIT \cite{liu2017unsupervised}.

Multi-model image translation which focuses on increasing the diversity of translated images rather than generating images in different domains, is an extension of two-domain image translation. An image can be converted to different styles or appearances in a same target domain, such as translating a sketch to several color photos. Recently, many studies are proposed to tackle this problem with adding random noise \cite{zhu2017toward} \cite{almahairi2018augmented} or random style features \cite{li2019asymmetric} to the translation. Zhu et. al \cite{zhu2017toward} use a low-dimensional latent vector which can be randomly sampled in testing to produce more diverse results, called BicycleGAN. Augmented CycleGAN proposed by Almahairi et.al \cite{almahairi2018augmented}, learns stochastic mappings which leverage auxiliary noise to capture multi-modal conditions. Li et al. \cite{li2019asymmetric} utilize an auxiliary variable to learn the extra information between two domains which have asymmetric information and then produce diverse target images. Huang et al. \cite{huang2018multimodal} directly extend UNIT \cite{liu2017unsupervised} to multi-modal scenarios called MUNIT, which encodes the images to a shared content space and combines a random domain-specialized style code for the generation.

\subsection{Multi-Domain Image-to-Image Translation}
Two-domain image-to-image translation models only handle an one-to-one mapping at a time. As the number of domains increases, the number of required mappings will exponentially increase. This makes them difficult to process the multi-domain image translation problems. Anoosheh et al. \cite{anoosheh2018combogan} who focus on reducing the modeling cost to linear complexity, divide the whole generator into $N$ encoders and $N$ decoders, and combine them to any pair to complete the translation between any two domains, called ComboGAN. A similarity idea can be found in Domain-Bank proposed by Hui et al. \cite{hui2018unsupervised}. They adopt a weight sharing constraint in the last few layers of encoders and the first few layers of decoders. Additional shared layers for the discriminators are also used to tie weights before the final output. Unlike the partition of the generator, Choi et al. \cite{choi2018stargan} use an auxiliary mask vector to label different domains and use only a single generator to translate multiple facial attributes, namely StarGAN. Following the same idea, Lin et al. \cite{lin2019relgan} use relative attributes in the label vector to transfer target facial attributes and preserve other non-target attributes, called RelGAN. Yu et.al \cite{yu2018singlegan} utilize domain code to explicitly control the different generation tasks. To perform image translation with scalability and diversity, Wang et al. \cite{wang2019sdit} condition the encoder with the target domain label and apply conditional instance normalization \cite{ulyanov2016instance} in their network architecture, called SDIT which provides a solution for both multi-modal and multi-domain image translation. A similar work is also done by xu et.al \cite{xu2019toward}. In general, existing strategies for multi-domain translations can be roughly classified as two types, splitting the generator or introducing an auxiliary label variable.


\begin{figure*}[t]
	\begin{center}
		\includegraphics[width=1\hsize]{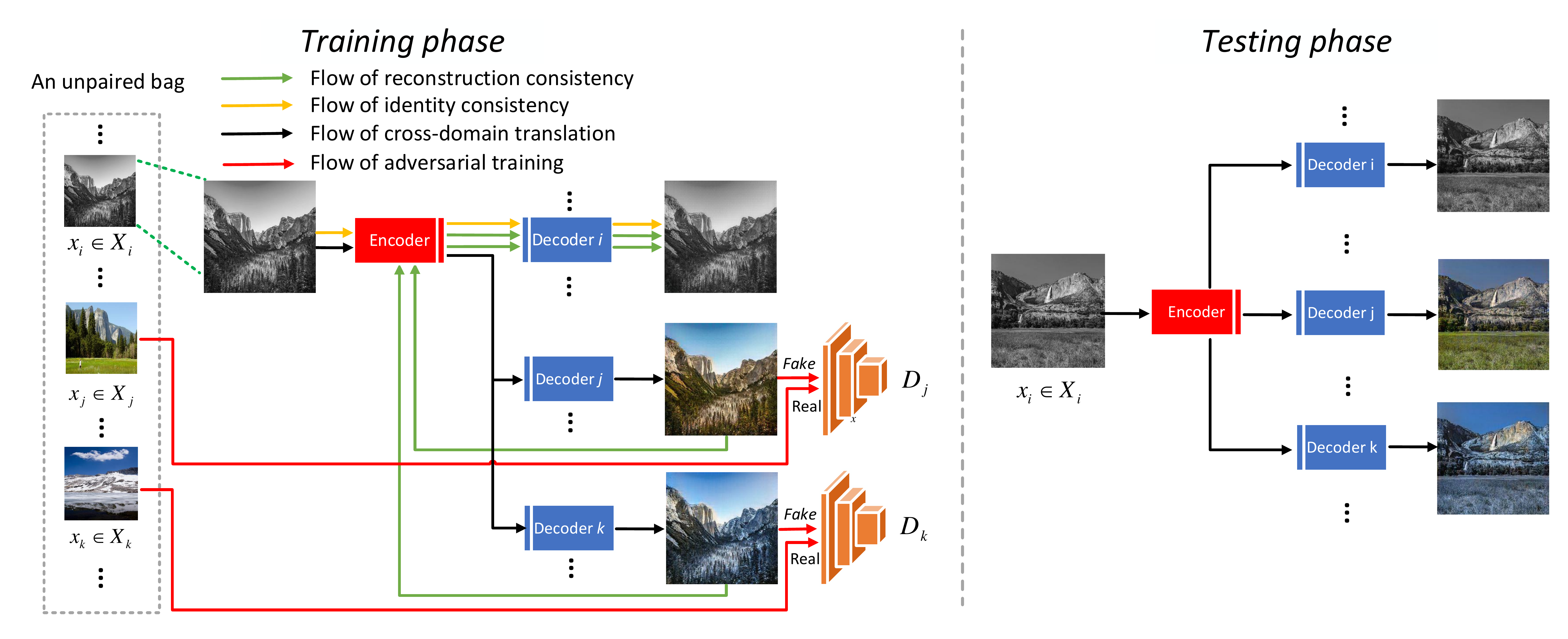}
	\end{center}
	\caption{The main process of training (left) and testing (right) in each domain for our method. During each training iteration, an unpaired bag consisting of images from different domains is fed to MDT. For an image of a certain domain, all decoders will participate in the generation and output $N$ fake images which are used to compute the adversarial loss and implement the two constraints during training. }
	\label{figs:procedure}
\end{figure*}	

\section{PROPOSED METHOD}\label{sec3}
Given different image domains represented by $\boldsymbol X_i, i \in  [1,N]$, the task of multi-domain image-to-image translation aims to find a representative mapping set $\{F_{ij}: \boldsymbol X_i \to \boldsymbol X_j\}, i, j \in  [1,N], i\neq j$. To simplify such $O(N^2)$ complexity of mapping set, we in turn focus on functions $\{F_i: \boldsymbol X \to \boldsymbol X_i\}$ that can map any source domains to a target domain. A shared encoder $E$ together with several domain-specialized decoders $G_i$ can meet this requirement. This network can embeds images from different domains to a shared latent space , thus reducing the interference of domain-specialized features from different source domains and making the generated images more consistent with the styles of the target domains.	We exploit GANs \cite{goodfellow2014generative} to conduct our scheme. A translation to domain $i$ is described as $G_i(E(\boldsymbol X))$, and can be rewritten as $G_i(\boldsymbol X)$ for the convenience of the following descriptions.

\subsection{Objective}
The overall objective of MDT consists of three parts:
\begin{align}
\mathcal{L}=\mathcal{L}_{adv}+\lambda_{rec}\mathcal{L}_{rec}+\lambda_{idt}\mathcal{L}_{idt}
\label{formula:total}
\end{align}
where $\mathcal{L}_{adv}, \mathcal{L}_{rec}, \mathcal{L}_{idt}$ represent the adversarial loss, two constraints of the reconstruction loss and the identity loss with their corresponding weights $\lambda_{rec}$ and $\lambda_{idt}$ to control their effects in training. Here below, details of $\mathcal{L}_{adv}, \mathcal{L}_{rec}, \mathcal{L}_{idt}$ are described.

\subsubsection{Adversarial loss} In unsupervised image translation with two image domains \cite{zhu2017unpaired} represented as $\boldsymbol X_i , \boldsymbol X_j, i,j \in \{1,2\}, i\neq j$, there are two mapping functions $F_{ij}: \boldsymbol X_i \to \boldsymbol X_j$. For each mapping, given $x_i  \in \boldsymbol X_i$ and $x_j  \in \boldsymbol X_j$ where $x_i,x_j$ are the images in their corresponding domains, the adversarial loss \cite{goodfellow2014generative} can be described as:

\begin{align}
\mathcal{L}_{ij}(G_j,D_j,\boldsymbol X_i,\boldsymbol X_j) &= \mathbb{E}_{x_j \sim P_{\boldsymbol X_j}}[f_1(D_j(x_j))] \nonumber \\
&+\mathbb{E}_{x_i \sim P_{\boldsymbol X_i}}[f_2(D_j(G_j(x_i))]
\label{formula:gan}
\end{align}		
where $D_j$, $G_j$ and $P_{\boldsymbol X_i}$ respectively represent the discriminator and generator for domain $j$ and the data distribution of $\boldsymbol X_i$, and the $f_1,f_2$ functions can be specified as $f_1(D)=log(D),f_2(D)=log(1-D)$ or other forms in different models \cite{gulrajani2017improved} \cite{isola2017image} \cite{zhu2017unpaired}. Following this objective, we can easily extend it to meet the requirements of translations among $N$ domains, where we only need to add up each item:
\begin{align}
\mathcal{L}_{adv}=\sum \limits_{i=1}^N  \sum \limits_{j=1}^N \mathcal{L}_{ij}, i \neq j
\label{formula:gan_total}
\end{align}

\subsubsection{Reconstruction loss} We generalize this loss from one-to-one mappings \cite{zhu2017unpaired} \cite{yi2017dualgan} \cite{kim2017learning}. It means that the input image should remain unchanged when it is successively processed by two generators with opposite input and output. This is an important idea which helps constrain the generation so that the translated images can retain more original contents and have the styles of the target domain. If there are two domains $\boldsymbol X_1,\boldsymbol X_2$ and two mapping functions $G_1: \boldsymbol X_2 \to \boldsymbol X_1, G_2: \boldsymbol X_1 \to \boldsymbol X_2$, the reconstruction constraint requires $G_1(G_2(\boldsymbol X_1))=\boldsymbol X_1, G_2(G_1(\boldsymbol X_2))=\boldsymbol X_2$. To measure the degree of approximation between $G_1(G_2(\boldsymbol X_1)$ and $\boldsymbol X_1$, $G_2(G_1(\boldsymbol X_2))$ and $\boldsymbol X_2$, we use the $L_1$ distance as the metric because of the less blurring for generating images \cite{isola2017image}. So the reconstruction loss for $\boldsymbol X_1, \boldsymbol X_2$  can be defined as $||G_1(G_2(\boldsymbol X_1))-\boldsymbol X_1||_1, ||G_2(G_1(\boldsymbol X_2))-\boldsymbol X_2||_1$.

For multi-domain scenario, the underlying intuition is that an input image from domain $i$, denoted as $x_i$, can be restored again after being transformed into all other domains except its source domain. Since $x_i$ will be transformed into $N-1$ domains, there are $N-1$ reconstruction terms by feeding each generated image $G_j(x_i)$ to the source domain generator $G_i$. Thus, the reconstruction loss for $x_i$ in domain $i$ can be defined as:
\begin{align}
\mathcal{L}_{rec}^i=\sum \limits_{j=1}^N||G_i(G_j(x_i))-x_i||_1, j \neq i
\end{align}
For the total loss of reconstruction, we only need to add up all the losses of each input domain:
\begin{align}
\mathcal{L}_{rec}=\sum \limits_{i=1}^N \mathcal{L}_{rec}^i
\end{align}

\subsubsection{Identity loss} In addition to reconstruction loss, we also propose another generalized loss as an extra constraint to improve the generation. The key is that when a domain generator processes an image from its own domain, it should allow the image to pass through without any changes. Thus the identity constraint means the invariance of generation in the generator's own domain. It helps enforce the generator to learn its own domain features. Following the same metric in the reconstruction loss, we define this constraint in domain $i$:
\begin{align}
\mathcal{L}^i_{idt}=||G_i(x_i)-x_i||_1
\end{align}
The total identity loss is thus:
\begin{align}
\mathcal{L}_{idt}=\sum \limits_{i=1}^N \mathcal{L}_{idt}^i
\end{align}

\subsection{Procedure}
\reffig{figs:procedure} illustrates the procedure of training and testing in MDT. During a training iteration, an unpaired bag consisting of images randomly selected from each domain is fed to the networks. Each image in this bag is processed by MDT to meet the two constraints and the adversarial loss. The colored lines respectively show the constraints of reconstruction (green) and identity (yellow), the processes of adversarial training (red), and the cross-domain translations (black). A common encoder (the red block) is used to embed any single input $x_i \in \boldsymbol X_i$ to the shared latent space, and a corresponding decoder  $G_j$ (the blue block) is provided to decode the embeddings to the desired domain $j$. For domain-specialized decoders and their corresponding discriminators, they are initially identical as each other, which means the labels of different domains are not special, so the only preparation for training data is to divide the images into different domains. For testing, we only need to feed an image from any domain to the generator, and then we will get the translated results in all domains, which is actually a many-to-many mapping. It seems unnecessary to generate an image in its source domain, but considering that MDT ignores the input domain label for practical utility, this is an inevitable result.

\subsection{Training Algorithm}
Here below is the training algorithm of MDT without considering the specific implementation or any algorithm used in each step. It does not separately train $N$ GANs models, but tied the encoder and all decoders together as a whole and accumulate all the losses obtained from different domains for backpropagation.
\begin{algorithm}
	\caption{MDT training procedure. The functions $f_1,f_2$ can be specialized into different GANs forms. Note that each generator always has a common encoding part with other generators, due to the shared encoder.}
	\label{alg}
	\begin{algorithmic}[1]
		\REQUIRE  $N$ image sets $\{\boldsymbol X_i\}^N_{i=1}$, $N$ generators $\{\boldsymbol G_i(\boldsymbol X;\omega_{g_i})\}^N_{i=1}$, and $N$ discriminators $\{\boldsymbol D_i(\boldsymbol X_i;\omega_{d_i})\}^N_{i=1}$.
		
		\STATE Randomly initialize all $\omega_{d_i},\omega_{g_i}, i \in [1,N]$. 	
		\FOR {number of training iterations}
		\STATE Randomly get an unpaired bag $\{x_i\}_{i=1}^N=\{x_i \in  \boldsymbol X_i,...,x_N \in \boldsymbol X_N \}$.
		\STATE Feed $\{G_1,...,G_N\}$ with $\{x_i\}_{i=1}^N$ to obtain $\{\hat{x}_{ij}\}_{i,j=1}^N=\{G_1(x_1),...,G_1(x_N),...,G_N(x_N)\}$.
		\FOR{$i=1,...,N$}
		\STATE update $\omega_{d_i}$ to minimize $\sum \limits_{j=1}^N 	[\mathbb{E}[f_1(D_i(x_i))]+\mathbb{E}[f_2(D_i(\hat{x}_{ij}))]]$.
		\ENDFOR
		\FOR{$i=1,...,N$}
		\STATE Compute $\mathcal{L}_{rec}^i=\sum \limits_{j=1}^N||G_i(\hat{x}_{ij})-x_i||_1, j \neq i$.
		\STATE Compute $\mathcal{L}_{idt}^i=||\hat{x}_{ii}-x_i||_1$.
		\STATE update $\omega_{g_i}$ to minimize $\sum \limits_{j=1}^N 	\mathbb{E}[f_2(D_i(G_i(x_j)))]+\lambda_{rec}\mathcal{L}_{rec}^i+\lambda_{idt}\mathcal{L}_{idt}^i$.
		\ENDFOR			
		\ENDFOR
	\end{algorithmic}
\end{algorithm}	 

\section{IMPLEMENTATION}\label{sec4}

\subsection{Network Architecture}
\begin{figure*}
	\begin{center}
		\includegraphics[width=0.9\hsize]{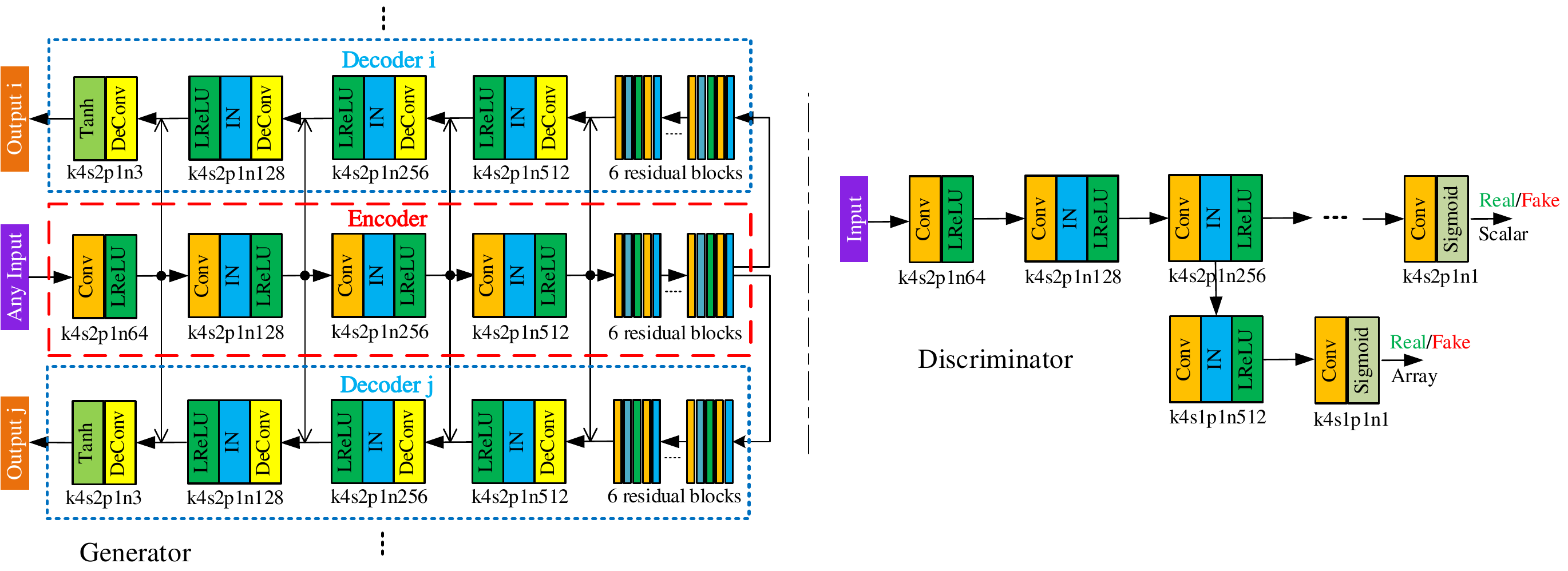}
	\end{center}
	\caption{The architecture of MDT with  the corresponding kernel size (k), stride (s), padding (p) and  number of feature maps (n). For the generator, the black lines are the flow of cross-domain translations, and the encoder or decoder is surrounded by a rectangle with red or blue dots. Here we show an example of two decoders labeled as $i$ and $j$.  For the discriminator, there are two outputs, an array and a scalar for the discrimination of the local and entire image.}
	\label{figs:generator}
\end{figure*}

\reffig{figs:generator} shows the architecture of our method.  For the generator, the encoder and decoders respectively contain several convolutional and deconvolutional layers with size-4, stride-2, and padding-1 to down-sample and up-sample, and have 6 residual blocks \cite{he2016deep} appended at the tail and header of them. The leaky rectified linear units (LReLU) with a slope of 0.2 is selected as the none linearity activation after instance normalization (IN) \cite{ulyanov2016instance}. To make full use of different feature maps in the generation, skip connections \cite{ronneberger2015u} are adopted between the same size features of encoder and decoders. For the discriminator, we selectively follow the conduction of Pix2Pix \cite{isola2017image}, using Markovian patchGAN \cite{li2016precomputed} architecture to discriminate whether $70\times70$ overlapping image patches are real or fake. To enhance the discriminator, the output of the original version is also retained. This means there are two outputs in a discriminator, an array and a scalar for the discrimination of the local and entire image.
\subsection{Training Details}
Instead of training generator (G) and discriminator (D) to minimize and max \refeq{formula:total}, we turn to respectively minimize the loss for G and D. Binary cross entropy (BCE) is used to measure the adversarial loss, and the real or fake image is labeled as 1 or 0 for the D. In detail,  with the measurement of BCE, \refeq{formula:gan} is equal to choose $f_1(D)=-logD, f_2(D)=-log(1-D)$ for training the D to minimize \refeq{formula:gan_total}. For the G, since the D only needs to regard fakes as reals, the measurement is equal to choosing $f_1(D)=0, f_2(D)=-logD$, and then the G is trained to minimize \refeq{formula:total}.

We use the Adam optimizer \cite{kingma2015adam} with momentum parameters $\beta_1 = 0.5$, $\beta_2 = 0.999$ and a batch size of 1, suggested in CycleGAN \cite{zhu2017unpaired}.  All the decoders are bundled with the encoder as a whole for backpropagation to update the weights, as are the discriminators.  All models are trained from scratch with a variable learning rate which is a constant value 0.0002 in the first half of epochs and then is linearly decayed to zero over the rest epochs. For all experiments, the images are resized to $256 \times 256$. We choose $\lambda_{rec}=\lambda_{idt}=10$ because we find this setting is helpful to balance the image quality of each translated domains.

\begin{figure*}
	\begin{center}
		\includegraphics[width=0.9\hsize]{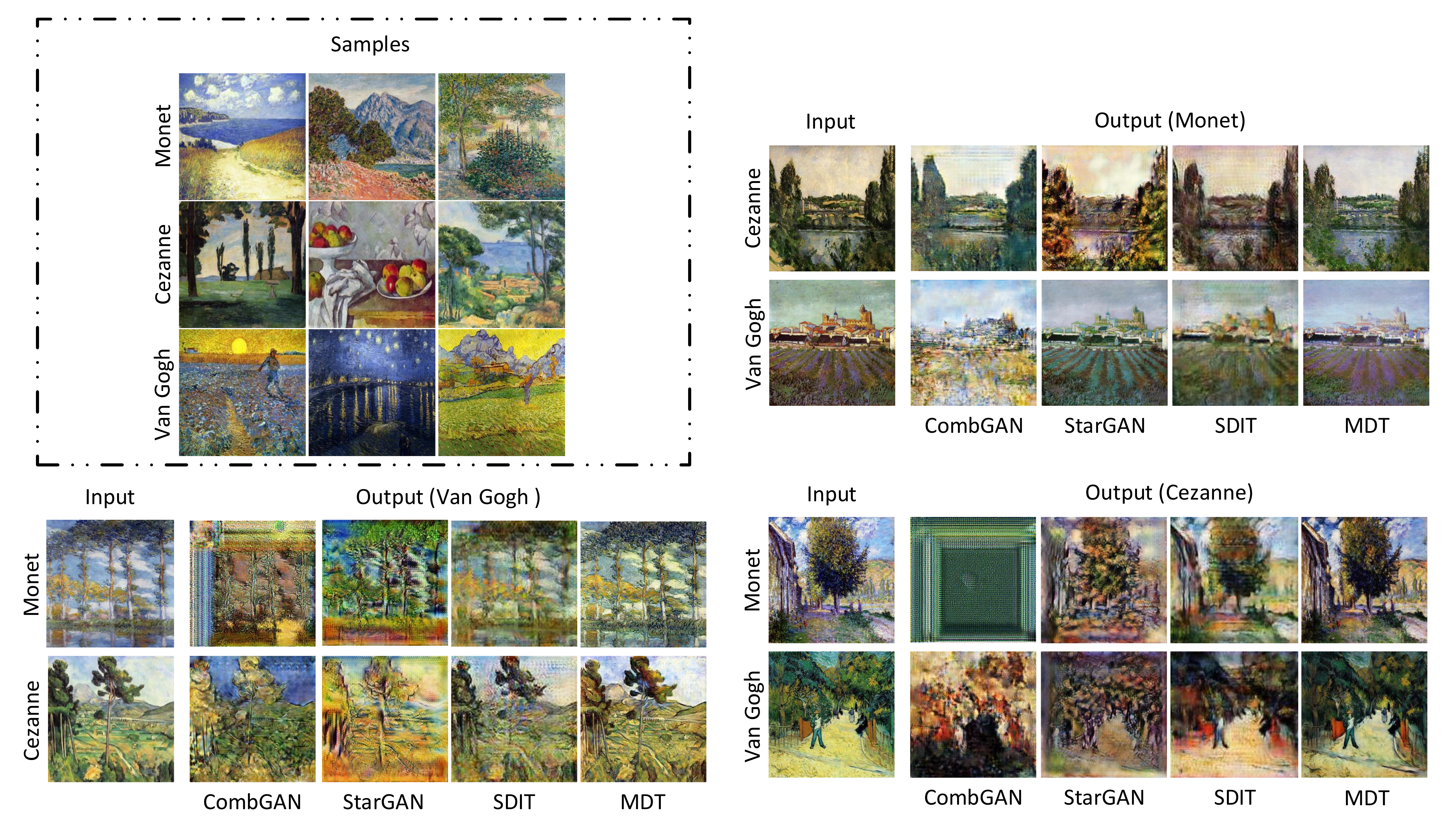}
	\end{center}
	\caption{Artistic style transfer among Monet, Cezanne, and Van Gogh. To make the comparison clear, the original samples are shown in left top, while other image patches respectively represent the translations from other styles (left column) to the target style (right columns) under different methods. }
	\label{figs:pic3}
\end{figure*}
\section{EXPERIMENTS}\label{sec5}
\subsection{Experimental Setting}\label{sec:datasets}
We adopt two types of unsupervised multi-domain methods as the baseline models. The one is CombGAN \cite{anoosheh2018combogan} , partitioning the generation. The others are StarGAN \cite{choi2018stargan} and SDIT \cite{wang2019sdit}, introducing an auxiliary label vector into the generation. All of them are representatives and have achieved state-of-the-art results. 

We conduct experiments on two types of datasets, images in each domain with or without ground truth in the other domains. All datasets contain multiple image domains. The one is Artistic Painting Styles, which includes five different unpaired images, dividing into four artistic painting styles and one scenic photo style, collected from Flickr and WikiArt \cite{deng2009imagenet}. Each domain has hundreds or thousands of images for training, as well as tens or hundreds for testing. The other is Multi-PIE \cite{gross2010multi}, which has more than 750, 000 face images under 15 poses, 20 illuminations and 6 expressions, taken from 337 subjects of different ages, genders, races and whether they wear glasses or not. In this database, we can divide the images into different domains, in which each image has ground truth in other domains. We choose different illuminations from session-1 of this database for face re-lighting synthesis. In each illumination domain, there are 249 face images corresponding to different subjects with the same pose and expression, divided into 150 and 99 for training and testing.

Here we evaluate all methods through the translations among three domains, because it is a typical multi-domain scenario with the minimum domain numbers. If a method performs unsatisfactory translations across three domains, it will also be ineffective in performing translations with more domain numbers. The style transfer among three different artistic styles and the face re-lighting under three illuminations namely normal, shadow, and dark, are selected as the tasks which represent two typical application scenarios of image-to-image translation, and respectively have large and small differences among their domains. Each task contains six sub cases, which are one-to-one mappings and are used for evaluation together with the whole task. We trained all models for 200 epochs with their default recommended super-parameters.

\subsection{Qualitative Evaluation}\label{sec:qualitative}

\reffig{figs:pic3} presents the translations among three different artistic painting styles, including the samples of original paintings (left top) for the convenience of clear comparison. CombGAN is failed and collapsed in some domain pairs and is also not good enough in its relatively successful results. For StarGAN and SDIT, they also have unsatisfactory translation results with obvious artifacts and ambiguities. Compared to these state-of-the-art methods, our approach transforms the images more realistically and accurately without obvious artifacts, and the transferred styles are more in line with the target styles.

\begin{figure}
	\begin{center}
		\includegraphics[width=0.9\hsize]{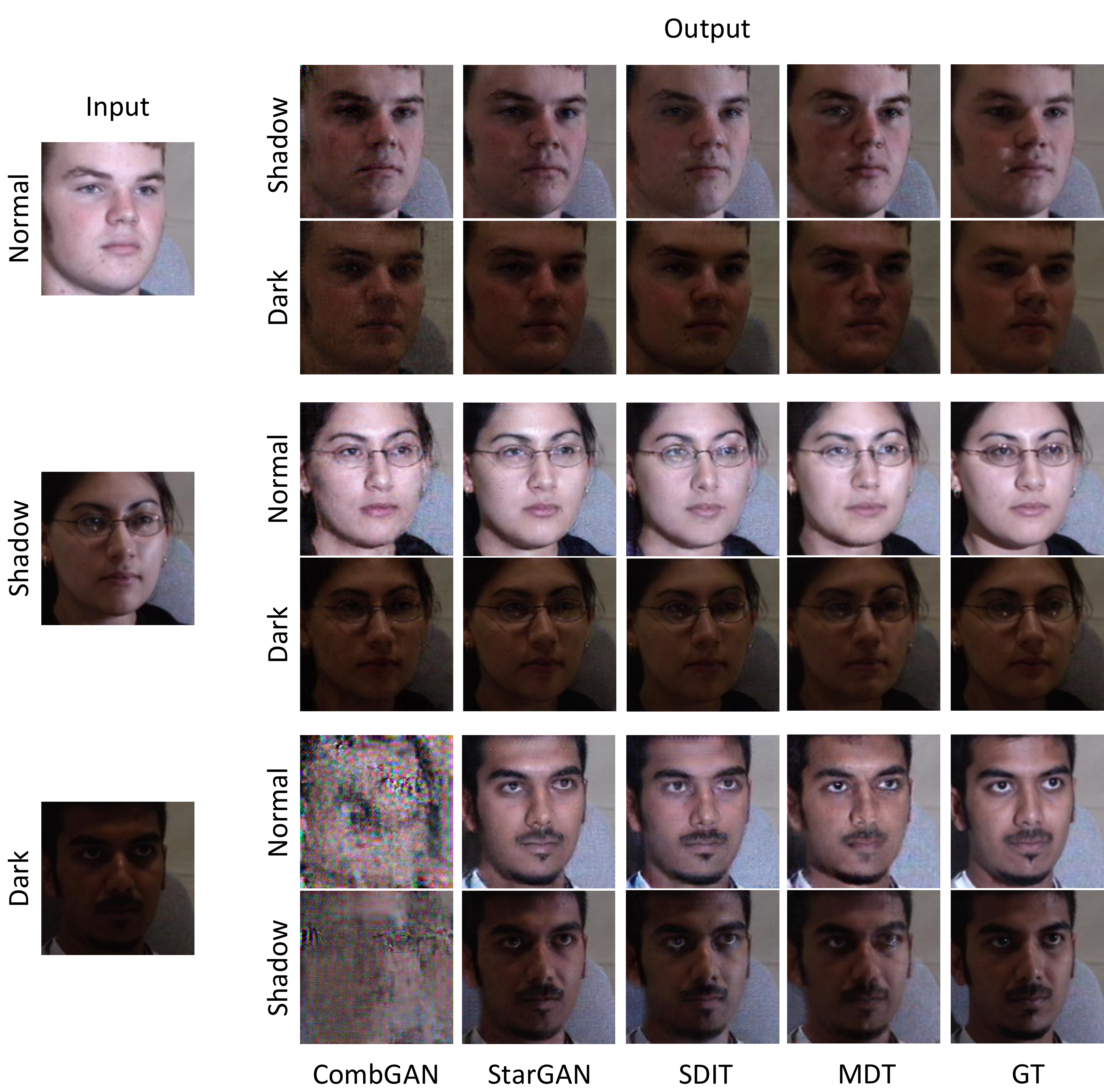}
	\end{center}
	\caption{Different methods for face re-lighting among three different illuminations of normal, shadow and dark. From left to right: input, CombGAN \cite{anoosheh2018combogan}, StarGAN \cite{choi2018stargan}, SDIT \cite{wang2019sdit}, MDT, and ground truth.}
	\label{figs:light3_view}
\end{figure}

The results of face re-lighting are shown in \reffig{figs:light3_view}. CombGAN is still unable to translate all sub cases well. The other three methods perform equally well, but according to the ground truth, StarGAN and MDT seem to retain more identity features after mapping.

\subsection{Quantitative Evaluation}\label{sec:quantitative}
\begin{table*}
	\renewcommand\arraystretch{0.98}
	\centering		
	\caption{FID and KID with standard deviations between the generated images and the real target images for the unpaired samples. For KID, we times 100 as the final results. For all scores, the lower is the better.}
	\resizebox{\textwidth}{!}{
		\begin{tabular}{c|cccc|cccc}
			\hline
			\textbf{Metric} & \multicolumn{4}{|c|}{FID} & \multicolumn{4}{|c}{KID$\times$100}         \\  \hline
			\diagbox{\textbf{Task}}{\textbf{Method}}&CombGAN&StarGAN&SDIT&MDT&CombGAN&StarGAN&SDIT&MDT\\ \hline
			
			Monet $\to$ Cezanne&
			418.29& 291.32 & 262.34 & \textbf{169.26} & 
			45.52$\pm$0.92 & 16.84$\pm$0.99 &13.91$\pm$1.02 &\textbf{4.89$\pm$1.24} \\

			Monet $\to$ Van Gogh&
			310.64 & 348.43 &306.94 & \textbf{231.19}& 
			25.80$\pm$2.34 & 23.42$\pm$2.18 &17.62$\pm$2.13& \textbf{11.08$\pm$1.76} \\

			Cezanne $\to$ Monet&
			185.03 & 252.66 & 283.93 &\textbf{151.65}& 
			10.56$\pm$0.82 & 10.94$\pm$1.18 & 15.17$\pm$1.26 & \textbf{4.03$\pm$0.66}\\
			
			Cezanne $\to$ Van Gogh&
			267.59 & 361.04 & 337.38 & \textbf{253.46} & 
			13.44$\pm$1.25 &22.44$\pm$1.21& 17.97$\pm$1.34 & \textbf{12.38$\pm$1.33} \\

			Van Gogh $\to$ Monet&
			236.46 & 259.07& 290.41 & \textbf{187.62} & 
			16.16$\pm$1.89 & 10.24$\pm$1.63 &  13.22$\pm$2.12 & \textbf{4.04$\pm$1.47} \\

			Van Gogh $\to$ Cezanne&
			323.75 & 320.06 & 296.17 & \textbf{197.21} & 
			20.81$\pm$1.76 & 18.32$\pm$1.24 & 13.31$\pm$1.35&\textbf{4.13$\pm$1.54}\\

			\hline
			Across all domains&
			290.29 & 305.43 & 296.20 & \textbf{198.40} & 
			22.05$\pm$1.50 & 17.03$\pm$1.41 & 15.20$\pm$1.54 & \textbf{6.76$\pm$1.33} \\
			\hline	
			
		\end{tabular}
	}
	
	\label{table:fid}	
\end{table*}
\begin{table*}
	\renewcommand\arraystretch{1}
	\centering		
	\caption{Classification accuracy and mean LIIPS on the style transfer task for each target domain. The higher is the better.}	
	\label{table:classify}	
	\resizebox{0.98\textwidth}{!}{
		\begin{tabular}{c|ccccc|ccccc}
			\hline
			\textbf{Metric} & \multicolumn{5}{|c|}{Classification Accuracy} & \multicolumn{5}{|c}{LPIPS}         \\  \hline
			\diagbox{\textbf{Domain}}{\textbf{Method}}&CombGAN&StarGAN&SDIT&MDT&Real image&CombGAN&StarGAN&SDIT&MDT&Real image\\ \hline
			
			Monet&
			\textbf{0.9669}&0.9583 &0.7250 & 0.8083&\slshape{0.9833}& 
			0.3872&0.3964&0.3402 & \textbf{0.4023}&\slshape{0.4186} \\ 
			
			Cezanne&
			0.0592&0.1908&0.1579&\textbf{0.8487}&\slshape{0.9775}& 
			0.2594&0.3226&0.3124& \textbf{0.3529}&\slshape{0.3737} \\ 
			
			Van Goph&
			0.5810&0.3571&0.1048 &\textbf{0.8381}& \slshape{0.9677} & 
			0.3929&0.3916 &0.3442&\textbf{0.4007} &\slshape{0.4052}\\  			
			
			\hline
			Across all domains&
			0.5357&0.5021 &0.3292&\textbf{0.8317}& \slshape{0.9762}& 
			0.3465&0.3702 &0.3323 &\textbf{0.3853}&   \slshape{0.3992}\\ 
			\hline

		\end{tabular}
	}
	
\end{table*}

Since the experiments are conducted on two types of datasets, one of which contains unpaired images, and the other in which all images have ground truth, we use corresponding metrics to quantitatively measure the results. 

\subsubsection{Evaluation of unpaired samples}
Fr\'{e}chet Inception Distance (FID) \cite{heusel2017gans} is usually used to evaluate the performance of GANs \cite{goodfellow2014generative}. It measures the distance between two samples which are real images and generated images, with the principle that if the distributions of two samples are more similar, FID value is more lower. Similar to FID, Kernel Inception Distance (KID) \cite{binkowski2018demystifying} is another metric to evaluate the generated images. Both of these two metrics utilize the Inception Network \cite{szegedy2015going} to obtain the image features to compute their own scores. \reftable{table:fid} illustrates the scores of FID and KID on the style transfer task. It is obvious that MDT has good performance in both the sub cases and the entire task against all the baseline methods, which demonstrates MDT improves the generation.  

To further evaluate whether the model correctly generates the target images, we employ the VGG-16 \cite{simonyan2015very} which is pre-trained on the ImageNet \cite{deng2009imagenet} database and fine-tuned on our dataset, to classify the generated images. To test whether there is a model collapse, we utilize Learned Perceptual Image Patch Similarity (LPIPS) \cite{zhang2018the} which measures the perceptual distance of output images, to reveal the diversity of different generations. The measurements are shown in \reftable{table:classify}. According to the classification scores, all baseline models almost fail in translating images to Cezanne domain, and they obviously are biased towards the Monet domain. For our method, it has a more balance performance in processing all domains. The measurements of LPIPS demonstrate that model collapse may possibly occur in CombGAN than other methods. For StarGAN and SDIT, the phenomenon of low classification accuracy with high LPIPS value in some domains implies that, these two methods are not effective enough in this style transfer task even if they output diverse images against model collapse. For our method, it stably translates the correct target images with good quality across all domains.

%
%


\subsubsection{Evaluation of paired samples}
\begin{figure}
	\begin{center}
		\includegraphics[width=0.9\hsize]{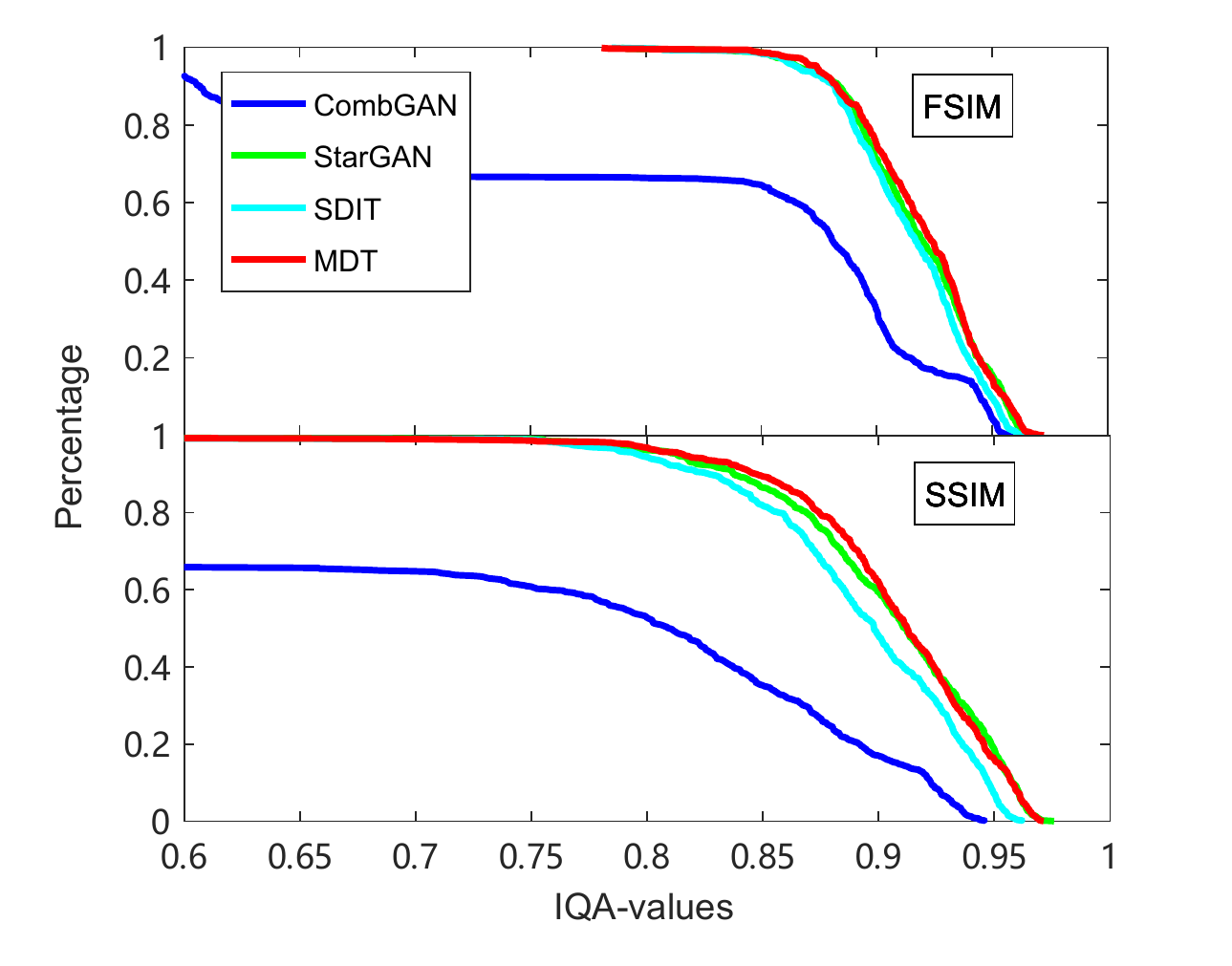}
	\end{center}
	\caption{FR-IQA values of all images generated in the entire task under different methods on the three-domain face-relighting task. Top and bottom: FSIM and SSIM values.}
	\label{figs:comparison_curve}
\end{figure}

In Multi-PIE \cite{gross2010multi} database, each image has its ground truth. Since some methods synthesized faces with good identity preservation, resulting in 100\% face recognition rate which can not provide a meaningful comparison, we turned to focus on Full Reference Image Quality Assessment (FR-IQA), and utilized two metrics of Feature Similarity Index (FSIM) \cite{zhang2011fsim} and Structural Similarity Index (SSIM) \cite{wang2004image} for evaluation. The principle of these two metrics is if two images are more similar, the measurement is more close to 1, which in our experiment represents the quality of generated images referenced to their ground truth. \reftable{table:fsim} lists the mean values and standard deviations of FSIM and SSIM for the six sub cases and the entire translation. It is obvious that CombGAN is still ineffective in translating some sub tasks, and StarGAN and MDT have the comparable performance. For more details of the overall assessment, we present the evaluation curves to illustrate the quality distribution of the generated images. As shown in \reffig{figs:comparison_curve}, the vertical axis indicates the percentage of images whose FR-IQA values are higher than the values on the horizontal axis. Our method almost has higher percentage on all IQA-values.

We do not list the evaluation results of classification accuracy and IPISP for this task, because almost all methods obtained 100\% accuracy, and the diversity of the original test sample is low making it meaningless to measure IPISP score. Following the work \cite{lin2020identity}, we use the cosine distance between the features of a fake face and its corresponding real one to measure whether a translated face is meet the target domain style and whether the original identity information is preserved. The ResNet-50 \cite{he2016deep} which is a high-performance off-the-shelf face recognition network and is pre-trained on the VGGFace2 database \cite{cao2018vggface2}, is employed to extract features for all real and fake images. The measurements of mean feature distance are shown in \reftable{table:cosine}. It is clear that CombGAN still has performance bias in some domains, and the other methods are successful in translations and have good preservation of the face identity information after mapping. On the whole, our method has the best generated results than the baseline models.

\begin{table*}
	\renewcommand\arraystretch{1.1}
	\centering		
	\caption{The mean and standard deviations of FSIM and SSIM under different methods in the face re-lighting task (including each sub cases). For all scores,the higher is the better.}	
	\resizebox{0.98\textwidth}{!}{
		\begin{tabular}{c|cccc|cccc}
			\hline
			\textbf{Metric} & \multicolumn{4}{|c|}{FSIM} & \multicolumn{4}{|c}{SSIM}         \\  \hline
			\diagbox{\textbf{Task}}{\textbf{Method}}&CombGAN&StarGAN&SDIT&MDT&CombGAN&StarGAN&SDIT&MDT\\ \hline
			
			normal $\to$ shadow&
			0.602$\pm$0.015 & 0.873$\pm$0.020 & 0.871$\pm$0.019 & \textbf{0.875$\pm$0.019} & 0.142$\pm$0.014 & 0.812$\pm$0.072 & 0.798$\pm$0.073 & \textbf{0.818$\pm$0.079} \\

			normal $\to$ dark&
			0.676$\pm$0.014 & 0.897$\pm$0.004 & 0.895$\pm$0.004 & \textbf{0.900$\pm$0.004} & 0.275$\pm$0.039 & 0.876$\pm$0.008 & 0.864$\pm$0.008 & \textbf{0.883$\pm$0.008} \\

			shadow $\to$ normal&
			0.865$\pm$0.016 & 0.912$\pm$0.004 & 0.909$\pm$0.005 & \textbf{0.916$\pm$0.005} & 0.749$\pm$0.081 & 0.901$\pm$0.007 & 0.887$\pm$0.006 & \textbf{0.903$\pm$0.005}\\
			
			shadow $\to$ dark&
			0.891$\pm$0.005 & 0.928$\pm$0.004 & 0.925$\pm$0.004 & \textbf{0.931$\pm$0.003 }& 0.833$\pm$0.013 & 0.922$\pm$0.006 & 0.909$\pm$0.007 &\textbf{0.922$\pm$0.005}\\

			dark $\to$ normal&
			0.907$\pm$0.007 & 0.941$\pm$0.004 & 0.936$\pm$0.004 & \textbf{0.941$\pm$0.003} & 0.879$\pm$0.012 & \textbf{0.943$\pm$0.006} & 0.931$\pm$0.005 & 0.940$\pm$0.006 \\

			dark $\to$ shadow&
			0.946$\pm$0.007 & \textbf{0.957$\pm$0.004} & 0.951$\pm$0.005 & 0.957$\pm$0.006 & 0.925$\pm$0.011 & \textbf{0.960$\pm$0.005} & 0.949$\pm$0.005 & 0.959$\pm$0.005 \\

			\hline
			Across all domains&
			0.815$\pm$0.129 & 0.918$\pm$0.029 &0.915$\pm$0.028 & \textbf{0.920$\pm$0.028} &
			0.634$\pm$0.310 & 0.902$\pm$0.057 &0.890$\pm$0.058 & \textbf{0.904$\pm$0.056}\\
			\hline

		\end{tabular}
	}
	
	\label{table:fsim}	
\end{table*}	
\begin{table}[!t]
	\renewcommand\arraystretch{1.1}
	\centering		
	\caption{Mean feature distance on the style transfer task for each domain. The lower is the better.}	
	\resizebox{0.45\textwidth}{!}{
		\begin{tabular}{c|cccc}
			\hline
			\diagbox{\textbf{Domain}}{\textbf{Method}}&CombGAN&StarGAN&SDIT&MDT\\ \hline
			
			Normal&
			0.5781&0.1977 &0.2125 &\textbf{0.1717}\\ 
			
			Shadow&
			0.5861 &\textbf{0.1575}&0.1622&0.1730\\ 
			
			Dark&
			0.1767&0.1428&0.1591&\textbf{0.1384}\\  			
			
			\hline
			Across all domains&
			0.4470&0.1660&0.1779&\textbf{0.1610}\\ 
			\hline

		\end{tabular}
	}
	
	\label{table:cosine}	
\end{table}		

\subsubsection{Evaluation summary}
In summary, CombGAN has failed in some sub cases of the two tasks, which may be due to the difficulty in balancing the training of each encoder and decoder. SDIT and StarGAN have achieved good visual effects in the face synthesis task, but they are not good enough in the style transfer task. This may be mainly because their single generator architecture has insufficient capacity to control the correct generation in the corresponding domain, when the differences among the target domains are large. For MDT, it is successful in the two tasks, and the translated results in each domain are almost superior to these of state-of-the-art methods. However, we observe that, for each method conducted in the two tasks, there are always some sub cases with relatively lower performance (e.g. ``Monet $\to$ Van Gogh'', ``normal $\to$ shadow''), which implies it is still a challenge to balance the training for each domain pair.

%

\begin{figure*}
	\centering
	\includegraphics[width=0.9\hsize]{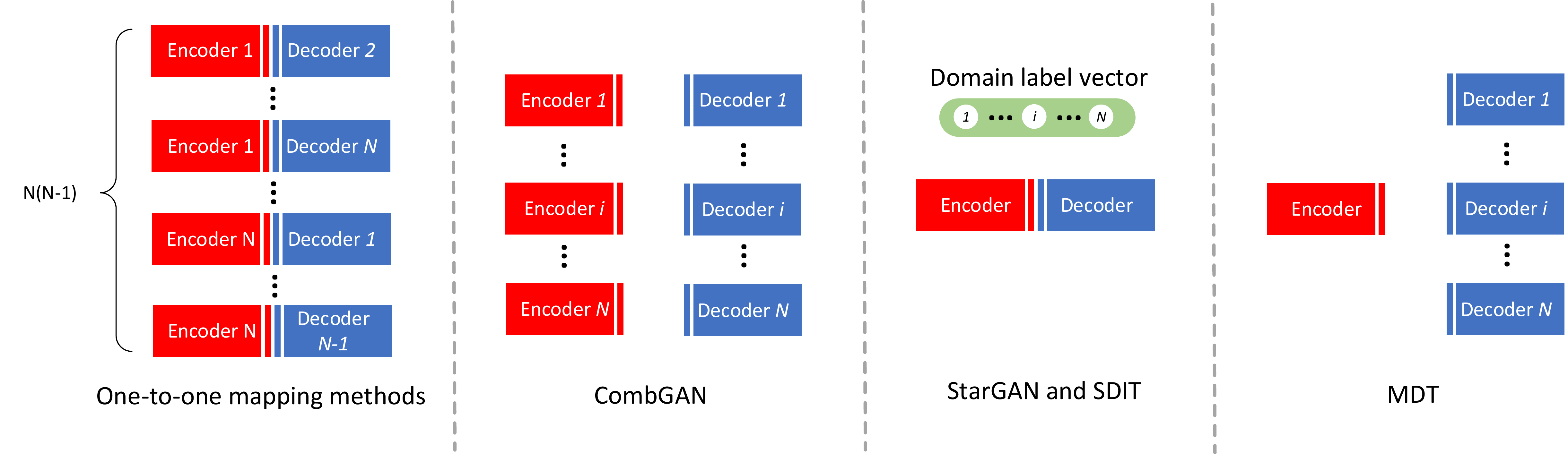}
	\caption{The schemes of different methods to efficiently process multi-domain image-to-image translation problems. For clear comparison, the traditional strategy of one-to-one mapping methods are also drawn in this figure.}
	\label{figs:scheme}
\end{figure*}	
\begin{table*}
	\renewcommand\arraystretch{1}
	\caption{Comparison for the attributes of different schemes used in the four methods. For testing complexity, $d$ and $e$ respectively are the complexities of the encoder and decoder. The attributes of model collapse, unbalance translation, and image quality are evaluated according to the experimental results.}	
	
	\centering		
	\resizebox{0.96\textwidth}{!}{
		\begin{tabular}{c|c|c|c|c}
			
			\hline
			\textbf{Scheme} & N encoders + N decoders & \multicolumn{2}{|c|}{encoder+decoder+label vector} & encoder + N decoders   \\  \hline
			
			\diagbox{\textbf{Attribute}}{\textbf{Method}}&CombGAN&StarGAN&SDIT&MDT\\ \hline
			
			Mapping type&
			one-to-many&\textbf{many-to-many}&\textbf{many-to-many}&\textbf{many-to-many}\\  				
			
			Network capacity&
			\textbf{best}&normal&normal&better\\  				
			
			Need input domain label&
			yes&\textbf{no}&\textbf{no}&\textbf{no}\\ \hline

			Training complexity&
			$O(2N)$&$\boldsymbol{O(1)} $&$\boldsymbol {O(1)}$&$O(N)$\\ 
			Testing complexity&
			$\boldsymbol {O(d+Ne)}$&$O(Nd+Ne)$ &$O(Nd+Ne)$&$\boldsymbol {O(d+Ne)}$\\ \hline

			Model collapse&
			yes&\textbf{no}&\textbf{no}&\textbf{no}\\ 
			
			Unbalance translation&
			yes&yes&yes&\textbf{no}\\ 
			
			Image quality&
			normal&better&better&\textbf{best}\\ 			
			\hline	
			
		\end{tabular}
	}
	\label{table:attribute}

\end{table*}	
\subsection{Analysis of Different Modeling Schemes}
Existing representative modeling schemes for multi-domain image-to-image translation are shown in \reffig{figs:scheme}. Compared to traditional one-to-one mappings, all the methods have improved the modeling efficiency. For StarGAN and SIDT, they have the same and the best efficient modeling scheme due to the introduction of a domain label vector to discriminate different domains. However, there is a trade-off between effectiveness and efficiency, and the most efficient translators do not mean the most effective translated results, as the evaluations mentioned in Section \ref{sec:quantitative} supported.

The detailed attributes of these schemes used in different methods are shown in \reftable{table:attribute}. The scheme used in CombGAN, splitting the generator into $N$ encoders and $N$ decoders to compose the corresponding processing for mappings among $N(N-1)$ domain pairs, provides enough capacity for different translations. It is efficient in its testing stage since an input image only needs to be encoded once, actually $O(d+Ne)$ where $d,e$ are respectively the complexity of its encoder and decoder. However, there are still too sub models (totally $2N$) making it difficult to balance the training of each domain pair, which may possibly cause failure or model collapse in the generation for some domain pairs. In addition, due to its one-to-many mapping property, it needs the source domain label to pick up the correct encoder, limiting its real applications.

For the scheme used in StarGAN and SDIT, it introduces a label vector into a single generator, and extremely improves the training efficiency, but it costs much in testing stage than the other schemes because its indivisible generator needs to encode the input image for each translation every time. Although it shows the ability against model collapse, it faces the difficulty of training domain classification for the label vector. If the label vector can not accurately discriminate different domains, it may lead to the same dilemma as CombGAN. In addition to this, the single generator may not have sufficient capacity to simultaneously handle translations across multiple domains with large differences but may be effective in processing translations with small differences, so StarGAN and SDIT are relatively successful in face synthesis but are not effective enough in style transfer. It is possible when there are conflicts for embedding requirements among different translations, a single mapping is unable to reconcile these conflicts, which can be avoided by using separate mappings. 

Compared with the scheme used in CombGAN, MDT needs to train for $N$ domain pairs, which reduces the difficulty of training balance among the $2N$ sub models. Furthermore, the shared encoder can reduce the interference of different domain-specialized information, making the $N$ decoders better complete their own translation. Compared with the scheme used in StarGAN and SDIT, the decoders also can be regarded as independent label vectors that do not need to train to classify different domains, thus avoiding the difficulty of controlling the correct generation. For the defects of ours compared to other schemes, they are the medium network capacity and medium training complexity, but it is effective and efficient enough for practical applications. Though the scheme quality also depends on the detailed model architecture, it is a key factor in determining the performance of a method. 

Since our scheme has only one encoder, we actually assume that all domains should have at least a same latent space. This assumption is reasonable, because we can always find an encoder to compress the data to a lower dimensional space to reduce the feature differences among domains. But if the shared embedding will cause a large loss of image content information, it may make the quality of the translated images degrade, or even make the translation fail. On the other hand, even if there is always an effective embedding space among target domains, we cannot assert that our scheme is stable for any number of domains, because as the domain number increases, there is less or even no information to share in embedding. We wonder if there will be a maximum number of translatable domains, over which MDT will not converge no matter what specific network structure or training technique is. Fortunately, MDT is sufficient to handle the number of domains involved in practical image-to-image translations.

In general, the typical advantages of the two schemes used in CombGAN, StarGAN and SDIT are respectively large network capacity, low modeling cost. While the typical disadvantages are respectively unbalance training, low network capacity. For our scheme, it has a trade-off between efficiency and effectiveness, making it more suitable for practical application. Certainly, the common difficulty for all of these solutions is how to further improve the unbalanced performance across all target domains.

\subsection{Analysis of the Objective}

In order to investigate the effectiveness of the proposed two generalized constraints, we isolate the items of reconstruction, identity and adversarial loss, and respectively train the networks to perform the three-domain face re-lighting task.

\begin{figure}
	\centering
	\includegraphics[width=0.9\hsize]{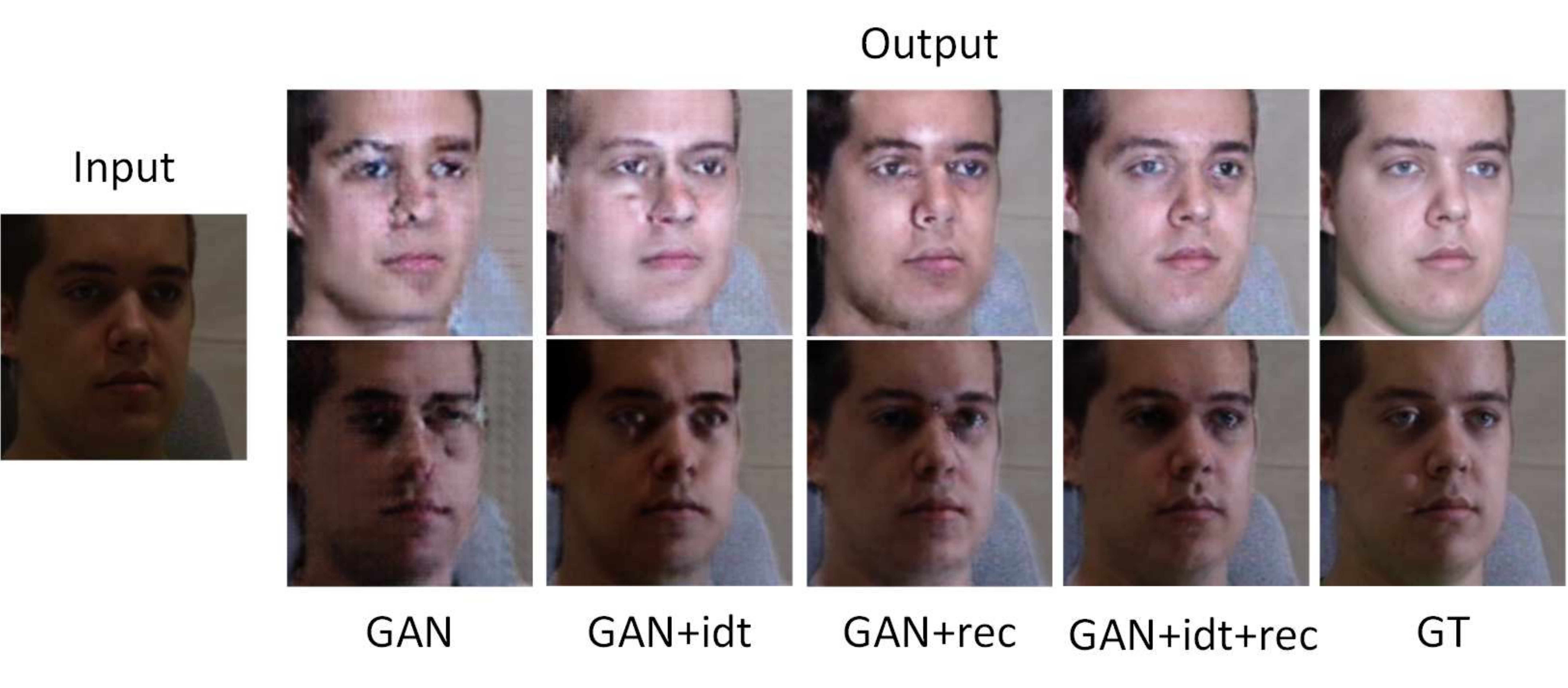}
	\caption{Different variants of our method for translations among three face illuminations. Here we only show an example of transferring dark (left) to normal (top) and shadow (bottom). From left to right: input, adversarial loss (GAN) alone, GAN with identity loss, GAN with reconstruction loss, our full loss and ground truth.}
	\label{figs:ablation_view}
\end{figure}

\begin{figure}
	\centering
	\includegraphics[width=0.9\hsize]{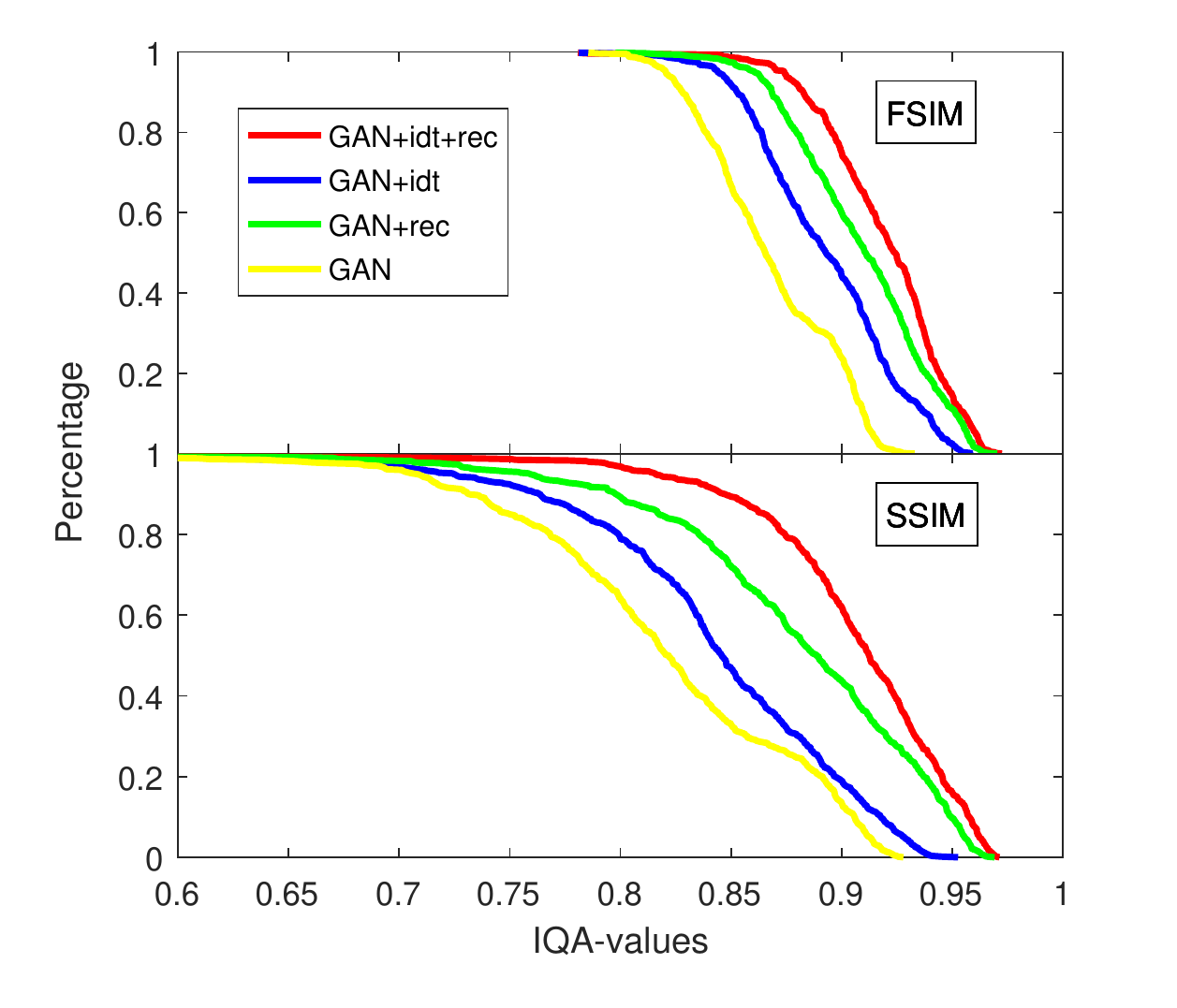}
	\caption{FR-IQA values of all generated images using various loss components in our method on the three-domain face re-lighting translation. Top and bottom: FSIM and SSIM values. }
	\label{figs:ablation_curve}
\end{figure}

\reffig{figs:ablation_view} shows an example of face re-lighting among dark, normal and shadow with using four different components to train. It is obvious that the image quality is improved with using reconstruction and identity consistency. We respectively measure the mean values and standard deviations of FSIM \cite{zhang2011fsim} and SSIM \cite{wang2004image} for different components as shown in \reftable{table:ablation}, where indicates the effectiveness of adding the reconstruction loss and identity loss in training. More details for analysis are draw in \reffig{figs:ablation_curve}. It can be seen that if these two constraints are not involved in training, the image generation quality of MDT will be substantially degraded.

In particular, compared to the identity loss, the reconstruction loss is more conducive to improve the generation, possibly because it has more processing and learning objectives than the other. In identity consistency, a decoder only focuses on learning the feature of its own domain through one input image, but in reconstruction constraint, it needs to use the learned domain features to restore $N-1$ fake images from all other domains.

\begin{table}
	\renewcommand\arraystretch{1}
	\centering
	\caption{The mean values and standard deviations of FSIM and SSIM on the three-domain face re-lighting task using variant loss in training MDT.}	
	\label{table:ablation}		
	\resizebox{0.4\textwidth}{!}{
		\begin{tabular}{c|cc}
			\hline
			\diagbox{\textbf{Loss}}{\textbf{Metric}}&
			FSIM &SSIM\\
			\hline
			
			GAN&
			0.87$\pm$0.03 & 0.82$\pm$0.07 \\
			
			GAN+idt&
			0.89$\pm$0.03 & 0.84$\pm$0.07\\
			
			GAN+rec&
			0.91$\pm$0.01 & 0.88$\pm$0.07\\
			
			GAN+rec+idt&
			0.92$\pm$0.02 & 0.90$\pm$0.05\\
			\hline	
		\end{tabular}
		
	}
\end{table}
\section{CONCLUSIONS}\label{sec6}
We propose an effective framework for unsupervised image-to-image translation across multiple domains, called MDT. It consists of a shared encoder and $N$ identical decoders, aiming to reduce the training complexity and the interference of the special source domain information. We also propose two general constraints extended from one-to-one mappings to meet the requirement of multi-domain scenario, which can significantly improve the quality of generated image. According to qualitative and quantitative evaluations, MDT performs favorably against the state-of-the-art multi-domain image translators \cite{anoosheh2018combogan} \cite{choi2018stargan} \cite{wang2019sdit} in both the entire task and each sub task, which suggests MDT provides an effective solution for image-to-image translations across multiple domains. In future work, we would like to extend MDT to handle other mixed domains, such as text, video or even audio.


\bibliographystyle{IEEEtrans}
\bibliography{ref}

\begin{thebibliography}{10}
\providecommand{\url}[1]{#1}
\csname url@samestyle\endcsname
\providecommand{\newblock}{\relax}
\providecommand{\bibinfo}[2]{#2}
\providecommand{\BIBentrySTDinterwordspacing}{\spaceskip=0pt\relax}
\providecommand{\BIBentryALTinterwordstretchfactor}{4}
\providecommand{\BIBentryALTinterwordspacing}{\spaceskip=\fontdimen2\font plus
\BIBentryALTinterwordstretchfactor\fontdimen3\font minus
  \fontdimen4\font\relax}
\providecommand{\BIBforeignlanguage}[2]{{%
\expandafter\ifx\csname l@#1\endcsname\relax
\typeout{** WARNING: IEEEtranS.bst: No hyphenation pattern has been}%
\typeout{** loaded for the language `#1'. Using the pattern for}%
\typeout{** the default language instead.}%
\else
\language=\csname l@#1\endcsname
\fi
#2}}
\providecommand{\BIBdecl}{\relax}
\BIBdecl

\bibitem{Abboud2003Expressive}
B.~Abboud, F.~Davoine, M.~Dang, and H.~Laboratory, ``Expressive face
  recognition and synthesis,'' \emph{Proceedings / CVPR, IEEE Computer Society
  Conference on Computer Vision and Pattern Recognition. IEEE Computer Society
  Conference on Computer Vision and Pattern Recognition}, 2003.

\bibitem{almahairi2018augmented}
A.~{Almahairi}, S.~{Rajeswar}, A.~{Sordoni}, P.~{Bachman}, and A.~{Courville},
  ``Augmented cyclegan: Learning many-to-many mappings from unpaired data,'' in
  \emph{ICML 2018: Thirty-fifth International Conference on Machine Learning},
  2018, pp. 195--204.

\bibitem{anoosheh2018combogan}
A.~{Anoosheh}, E.~{Agustsson}, R.~{Timofte}, and L.~V. {Gool}, ``Combogan:
  Unrestrained scalability for image domain translation,'' in \emph{2018
  IEEE/CVF Conference on Computer Vision and Pattern Recognition Workshops
  (CVPRW)}, 2018, pp. 783--790.

\bibitem{bhatt2012memetically}
H.~S. Bhatt, S.~Bharadwaj, R.~Singh, and M.~Vatsa, ``Memetically optimized
  mcwld for matching sketches with digital face images,'' \emph{IEEE
  Transactions on Information Forensics and Security}, vol.~7, no.~5, pp.
  1522--1535, 2012.

\bibitem{binkowski2018demystifying}
M.~{Bińkowski}, D.~J. {Sutherland}, M.~{Arbel}, and A.~{Gretton},
  ``Demystifying mmd gans,'' in \emph{ICLR 2018 : International Conference on
  Learning Representations 2018}, 2018.

\bibitem{cao2018vggface2}
Q.~{Cao}, L.~{Shen}, W.~{Xie}, O.~M. {Parkhi}, and A.~{Zisserman}, ``Vggface2:
  A dataset for recognising faces across pose and age,'' in \emph{2018 13th
  IEEE International Conference on Automatic Face and Gesture Recognition (FG
  2018)}, 2018, pp. 67--74.

\bibitem{choi2018stargan}
Y.~{Choi}, M.~{Choi}, M.~{Kim}, J.-W. {Ha}, S.~{Kim}, and J.~{Choo}, ``Stargan:
  Unified generative adversarial networks for multi-domain image-to-image
  translation,'' in \emph{2018 IEEE/CVF Conference on Computer Vision and
  Pattern Recognition}, 2018, pp. 8789--8797.

\bibitem{deng2009imagenet}
J.~{Deng}, W.~{Dong}, R.~{Socher}, L.-J. {Li}, K.~{Li}, and L.~{Fei-Fei},
  ``Imagenet: A large-scale hierarchical image database,'' in \emph{2009 IEEE
  Conference on Computer Vision and Pattern Recognition}, 2009, pp. 248--255.

\bibitem{goodfellow2014generative}
I.~J. {Goodfellow}, J.~{Pouget-Abadie}, M.~{Mirza}, B.~{Xu}, D.~{Warde-Farley},
  S.~{Ozair}, A.~C. {Courville}, and Y.~{Bengio}, ``Generative adversarial
  nets,'' in \emph{Advances in Neural Information Processing Systems 27}, 2014,
  pp. 2672--2680.

\bibitem{gross2010multi}
R.~{Gross}, I.~{Matthews}, J.~{Cohn}, T.~{Kanade}, and S.~{Baker},
  ``Multi-pie,'' \emph{Image and Vision Computing}, vol.~28, no.~5, pp.
  807--813, 2010.

\bibitem{gulrajani2017improved}
I.~{Gulrajani}, F.~{Ahmed}, M.~{Arjovsky}, V.~{Dumoulin}, and A.~C.
  {Courville}, ``Improved training of wasserstein gans,'' in \emph{Advances in
  Neural Information Processing Systems}, 2017, pp. 5767--5777.

\bibitem{he2016deep}
K.~{He}, X.~{Zhang}, S.~{Ren}, and J.~{Sun}, ``Deep residual learning for image
  recognition,'' in \emph{2016 IEEE Conference on Computer Vision and Pattern
  Recognition (CVPR)}, 2016, pp. 770--778.

\bibitem{heusel2017gans}
M.~Heusel, H.~Ramsauer, T.~Unterthiner, B.~Nessler, and S.~Hochreiter, ``Gans
  trained by a two time-scale update rule converge to a local nash
  equilibrium,'' in \emph{Advances in Neural Information Processing Systems},
  2017, pp. 6626--6637.

\bibitem{huang2018multimodal}
X.~{Huang}, M.-Y. {Liu}, S.~J. {Belongie}, and J.~{Kautz}, ``Multimodal
  unsupervised image-to-image translation,'' in \emph{Proceedings of the
  European Conference on Computer Vision (ECCV)}, 2018, pp. 179--196.

\bibitem{hui2018unsupervised}
L.~{Hui}, X.~{Li}, J.~{Chen}, H.~{He}, and J.~{Yang}, ``Unsupervised
  multi-domain image translation with domain-specific encoders/decoders,'' in
  \emph{2018 24th International Conference on Pattern Recognition (ICPR)},
  2018, pp. 2044--2049.

\bibitem{isola2017image}
P.~{Isola}, J.-Y. {Zhu}, T.~{Zhou}, and A.~A. {Efros}, ``Image-to-image
  translation with conditional adversarial networks,'' in \emph{2017 IEEE
  Conference on Computer Vision and Pattern Recognition (CVPR)}, 2017, pp.
  5967--5976.

\bibitem{johnson2016perceptual}
J.~{Johnson}, A.~{Alahi}, and L.~{Fei-Fei}, ``Perceptual losses for real-time
  style transfer and super-resolution,'' in \emph{European Conference on
  Computer Vision (ECCV)}, 2016, pp. 694--711.

\bibitem{kazemi2018facial}
H.~{Kazemi}, M.~{Iranmanesh}, A.~{Dabouei}, S.~{Soleymani}, and N.~M.
  {Nasrabadi}, ``Facial attributes guided deep sketch-to-photo synthesis,'' in
  \emph{2018 IEEE Winter Applications of Computer Vision Workshops (WACVW)},
  2018, pp. 1--8.

\bibitem{kim2017learning}
T.~{Kim}, M.~{Cha}, H.~{Kim}, J.~K. {Lee}, and J.~{Kim}, ``Learning to discover
  cross-domain relations with generative adversarial networks,'' in
  \emph{ICML'17 Proceedings of the 34th International Conference on Machine
  Learning - Volume 70}, 2017, pp. 1857--1865.

\bibitem{kingma2015adam}
D.~P. {Kingma} and J.~L. {Ba}, ``Adam: A method for stochastic optimization,''
  \emph{International Conference on Learning Representations}, 2015.

\bibitem{ledig2017photo}
C.~{Ledig}, L.~{Theis}, F.~{Huszar}, J.~{Caballero}, A.~{Cunningham},
  A.~{Acosta}, A.~P. {Aitken}, A.~{Tejani}, J.~{Totz}, Z.~{Wang}, and W.~{Shi},
  ``Photo-realistic single image super-resolution using a generative
  adversarial network,'' in \emph{2017 IEEE Conference on Computer Vision and
  Pattern Recognition (CVPR)}, 2017, pp. 105--114.

\bibitem{li2016precomputed}
C.~{Li} and M.~{Wand}, ``Precomputed real-time texture synthesis with markovian
  generative adversarial networks,'' in \emph{European Conference on Computer
  Vision}, 2016, pp. 702--716.

\bibitem{li2019asymmetric}
Y.~{Li}, S.~{Tang}, R.~{Zhang}, Y.~{Zhang}, J.~{Li}, and S.~{Yan}, ``Asymmetric
  gan for unpaired image-to-image translation,'' \emph{IEEE Transactions on
  Image Processing}, vol.~28, no.~12, pp. 5881--5896, 2019.

\bibitem{lin2020identity}
Y.~{Lin}, S.~{Ling}, K.~{Fu}, and P.~{Cheng}, ``An identity-preserved model for
  face sketch-photo synthesis,'' \emph{IEEE Signal Processing Letters},
  vol.~27, pp. 1095--1099, 2020.

\bibitem{lin2019relgan}
Y.-J. {Lin}, P.-W. {Wu}, C.-H. {Chang}, E.~{Chang}, and S.-W. {Liao}, ``Relgan:
  Multi-domain image-to-image translation via relative attributes,'' in
  \emph{2019 IEEE/CVF International Conference on Computer Vision (ICCV)},
  2019, pp. 5913--5921.

\bibitem{liu2017unsupervised}
M.-Y. {Liu}, T.~{Breuel}, and J.~{Kautz}, ``Unsupervised image-to-image
  translation networks,'' in \emph{Advances in Neural Information Processing
  Systems}, 2017, pp. 700--708.

\bibitem{liu2016coupled}
M.-Y. {Liu} and O.~{Tuzel}, ``Coupled generative adversarial networks,''
  \emph{arXiv preprint arXiv:1606.07536}, 2016.

\bibitem{mirza2014conditional}
M.~{Mirza} and S.~{Osindero}, ``Conditional generative adversarial nets,''
  \emph{arXiv preprint arXiv:1411.1784}, 2014.

\bibitem{papandreou2015weakly}
G.~{Papandreou}, L.-C. {Chen}, K.~P. {Murphy}, and A.~L. {Yuille}, ``Weakly-and
  semi-supervised learning of a deep convolutional network for semantic image
  segmentation,'' in \emph{2015 IEEE International Conference on Computer
  Vision (ICCV)}, 2015, pp. 1742--1750.

\bibitem{pathak2016context}
D.~{Pathak}, P.~{Krähenbühl}, J.~{Donahue}, T.~{Darrell}, and A.~A. {Efros},
  ``Context encoders: Feature learning by inpainting,'' in \emph{2016 IEEE
  Conference on Computer Vision and Pattern Recognition (CVPR)}, 2016, pp.
  2536--2544.

\bibitem{Tylecek13}
R.~{\v S}. Radim~Tyle{\v c}ek, ``Spatial pattern templates for recognition of
  objects with regular structure,'' in \emph{Proc. GCPR}, Saarbrucken, Germany,
  2013.

\bibitem{ronneberger2015u}
O.~{Ronneberger}, P.~{Fischer}, and T.~{Brox}, ``U-net: Convolutional networks
  for biomedical image segmentation,'' \emph{Medical Image Computing and
  Computer Assisted Intervention}, pp. 234--241, 2015.

\bibitem{shen2019one}
Z.~{Shen}, S.~K. {Zhou}, Y.~{Chen}, B.~{Georgescu}, X.~{Liu}, and T.~S.
  {Huang}, ``One-to-one mapping for unpaired image-to-image translation,''
  \emph{The IEEE Winter Conference on Applications of Computer Vision}, pp.
  1170--1179, 2019.

\bibitem{simonyan2015very}
K.~{Simonyan} and A.~{Zisserman}, ``Very deep convolutional networks for
  large-scale image recognition,'' in \emph{ICLR 2015 : International
  Conference on Learning Representations 2015}, 2015.

\bibitem{szegedy2015going}
C.~{Szegedy}, W.~{Liu}, Y.~{Jia}, P.~{Sermanet}, S.~{Reed}, D.~{Anguelov},
  D.~{Erhan}, V.~{Vanhoucke}, and A.~{Rabinovich}, ``Going deeper with
  convolutions,'' in \emph{2015 IEEE Conference on Computer Vision and Pattern
  Recognition (CVPR)}, 2015, pp. 1--9.

\bibitem{taigman2017unsupervised}
Y.~{Taigman}, A.~{Polyak}, and L.~{Wolf}, ``Unsupervised cross-domain image
  generation,'' in \emph{ICLR 2017 : International Conference on Learning
  Representations 2017}, 2017.

\bibitem{tang2018dual}
H.~{Tang}, D.~{Xu}, W.~{Wang}, Y.~{Yan}, and N.~{Sebe}, ``Dual generator
  generative adversarial networks for multi-domain image-to-image
  translation.'' in \emph{Asian Conference on Computer Vision}, 2018, pp.
  3--21.

\bibitem{ulyanov2016instance}
D.~{Ulyanov}, A.~{Vedaldi}, and V.~S. {Lempitsky}, ``Instance normalization:
  The missing ingredient for fast stylization.'' \emph{arXiv preprint
  arXiv:1607.08022}, 2016.

\bibitem{wang2009face}
X.~{Wang} and X.~{Tang}, ``Face photo-sketch synthesis and recognition,''
  \emph{IEEE Transactions on Pattern Analysis and Machine Intelligence},
  vol.~31, no.~11, pp. 1955--1967, 2009.

\bibitem{wang2019sdit}
Y.~{Wang}, A.~{Gonzalez-Garcia}, J.~van~de {Weijer}, and L.~{Herranz}, ``Sdit:
  Scalable and diverse cross-domain image translation,'' in \emph{Proceedings
  of the 27th ACM International Conference on Multimedia}, 2019, pp.
  1267--1276.

\bibitem{wang2004image}
Z.~{Wang}, A.~{Bovik}, H.~{Sheikh}, and E.~{Simoncelli}, ``Image quality
  assessment: From error visibility to structural similarity,'' \emph{IEEE
  Transactions on Image Processing}, vol.~13, no.~4, pp. 600--612, 2004.

\bibitem{Xiao2018Pattern}
B.~Xiao, J.~Zhou, and A.~Robles-Kelly, ``Pattern recognition for high
  performance imaging,'' \emph{Pattern Recognition}, vol.~82, 2018.

\bibitem{xu2019toward}
W.~{Xu}, K.~{Shawn}, and G.~{Wang}, ``Toward learning a unified many-to-many
  mapping for diverse image translation,'' \emph{Pattern Recognition}, vol.~93,
  pp. 570--580, 2019.

\bibitem{yi2017dualgan}
Z.~{Yi}, H.~{Zhang}, P.~{Tan}, and M.~{Gong}, ``Dualgan: Unsupervised dual
  learning for image-to-image translation,'' in \emph{2017 IEEE International
  Conference on Computer Vision (ICCV)}, 2017, pp. 2868--2876.

\bibitem{yu2018singlegan}
X.~{Yu}, X.~{Cai}, Z.~{Ying}, T.~H. {Li}, and G.~{Li}, ``Singlegan:
  Image-to-image translation by a single-generator network using multiple
  generative adversarial learning.'' in \emph{Asian Conference on Computer
  Vision}, 2018, pp. 341--356.

\bibitem{zhang2011fsim}
L.~{Zhang}, L.~{Zhang}, X.~{Mou}, and D.~{Zhang}, ``Fsim: A feature similarity
  index for image quality assessment,'' \emph{IEEE Transactions on Image
  Processing}, vol.~20, no.~8, pp. 2378--2386, 2011.

\bibitem{zhang2016colorful}
R.~{Zhang}, P.~{Isola}, and A.~A. {Efros}, ``Colorful image colorization,'' in
  \emph{European Conference on Computer Vision}, 2016, pp. 649--666.

\bibitem{zhang2018the}
R.~{Zhang}, P.~{Isola}, A.~A. {Efros}, E.~{Shechtman}, and O.~{Wang}, ``The
  unreasonable effectiveness of deep features as a perceptual metric,'' in
  \emph{2018 IEEE/CVF Conference on Computer Vision and Pattern Recognition},
  2018, pp. 586--595.

\bibitem{zhang2011coupled}
W.~{Zhang}, X.~{Wang}, and X.~{Tang}, ``Coupled information-theoretic encoding
  for face photo-sketch recognition,'' in \emph{CVPR 2011: IEEE Conference on
  Computer Vision and Pattern Recognition 2011}, 2011, pp. 513--520.

\bibitem{zhu2017unpaired}
J.-Y. {Zhu}, T.~{Park}, P.~{Isola}, and A.~A. {Efros}, ``Unpaired
  image-to-image translation using cycle-consistent adversarial networks,'' in
  \emph{2017 IEEE International Conference on Computer Vision (ICCV)}, 2017,
  pp. 2242--2251.

\bibitem{zhu2017toward}
J.-Y. {Zhu}, R.~{Zhang}, D.~{Pathak}, T.~{Darrell}, A.~A. {Efros}, O.~{Wang},
  and E.~{Shechtman}, ``Toward multimodal image-to-image translation,'' in
  \emph{Advances in Neural Information Processing Systems}, 2017, pp. 465--476.

\end{thebibliography}

\clearpage

\section*{SUPPLEMENTARY MATERIAL}\label{sec7}

\begin{figure}[!t]
	\centering
	\includegraphics[width=0.9\hsize]{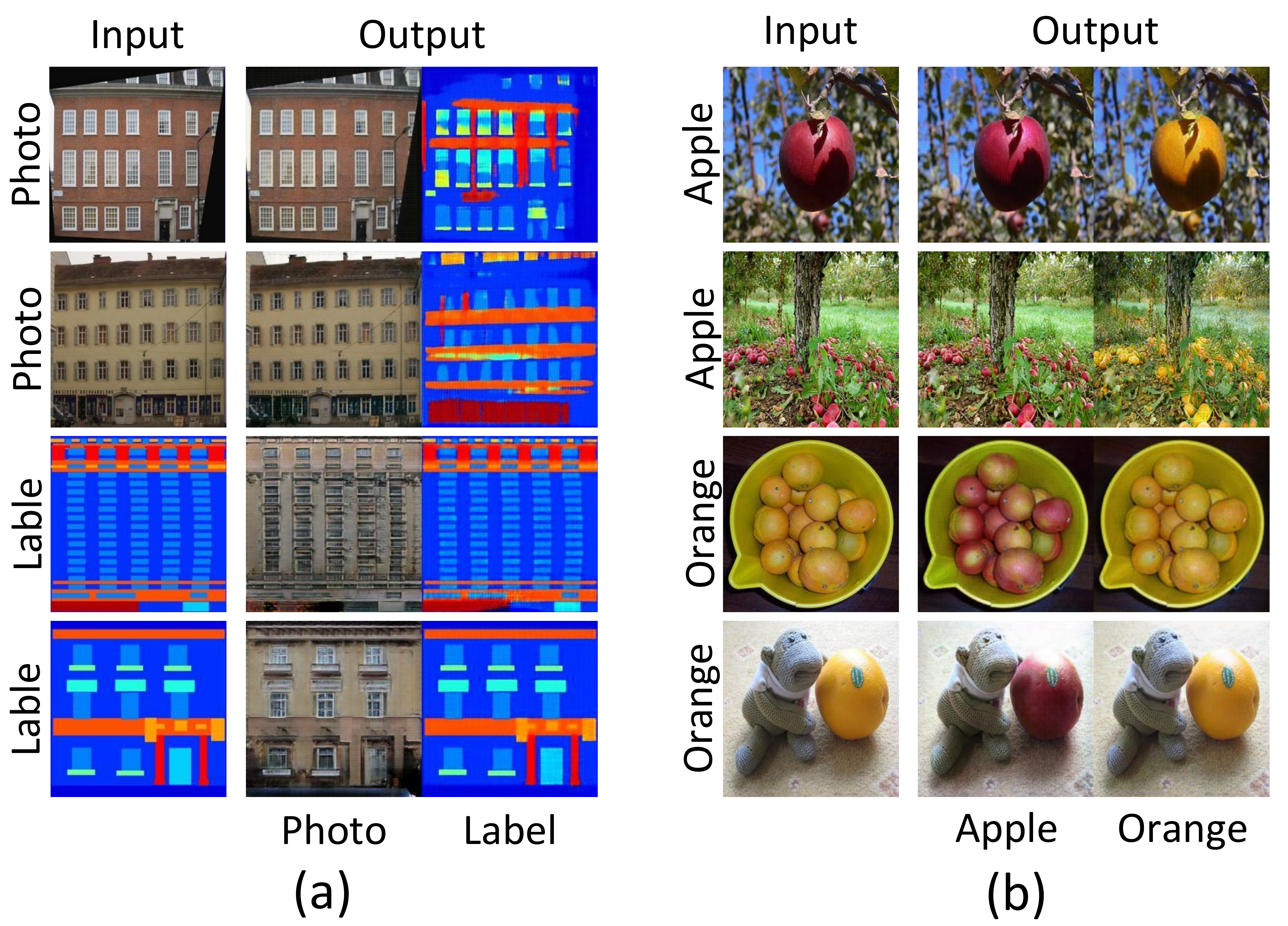}
	\caption{Results of two-domain translations: (a) photo  $\leftrightarrow$  semantic label, and (b) apple $\leftrightarrow$ orange. Using MDT, there is no contamination on non-target objects in generated images of apples and oranges. }
	\label{figs:domain2}
\end{figure}

\begin{figure}
	\centering
	\includegraphics[width=0.9\hsize]{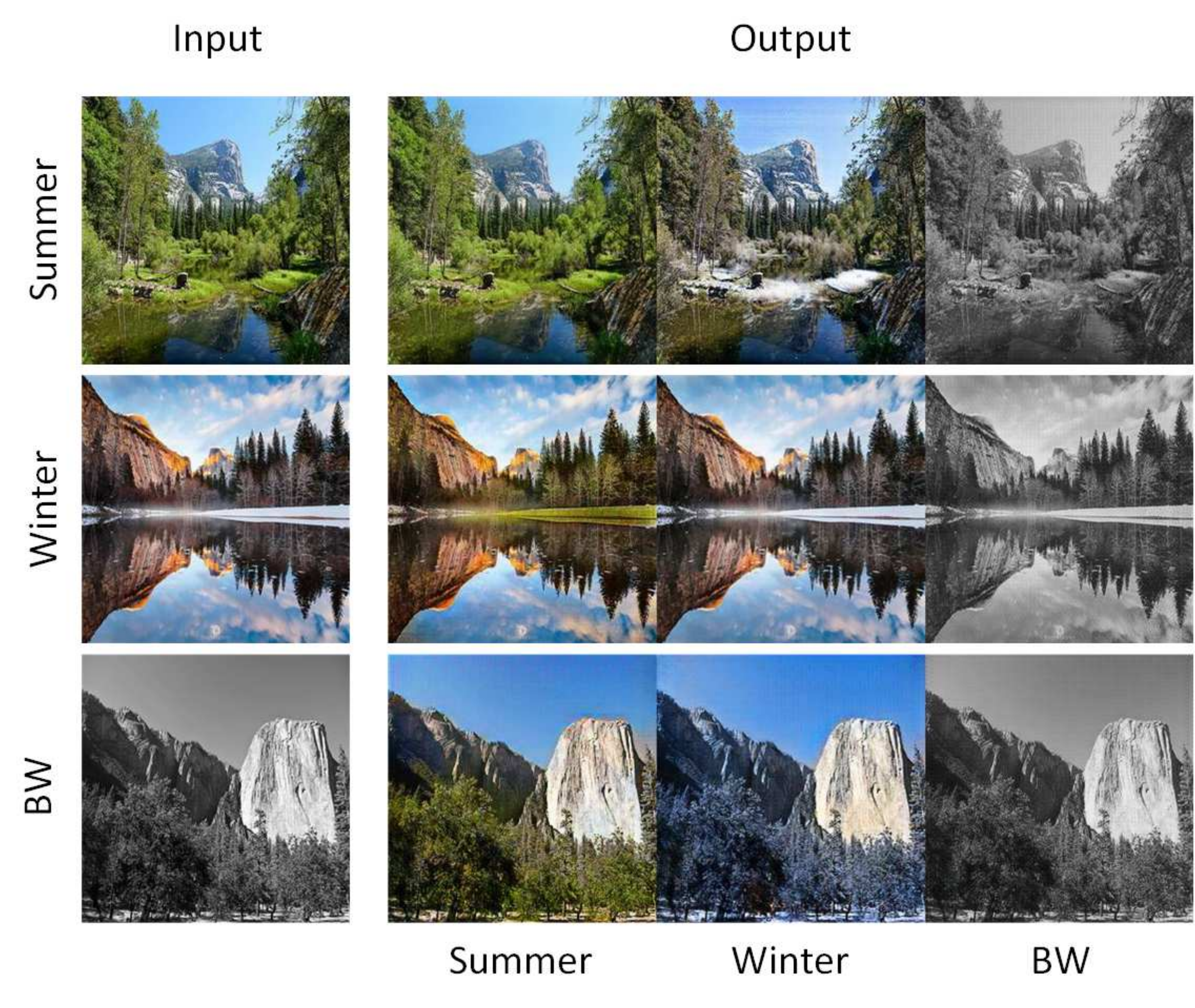}
	\caption{Example of translation among summer, winter and BW photos. }
	\label{figs:domain3}
\end{figure}

\begin{figure}
	\centering
	\includegraphics[width=0.9\hsize]{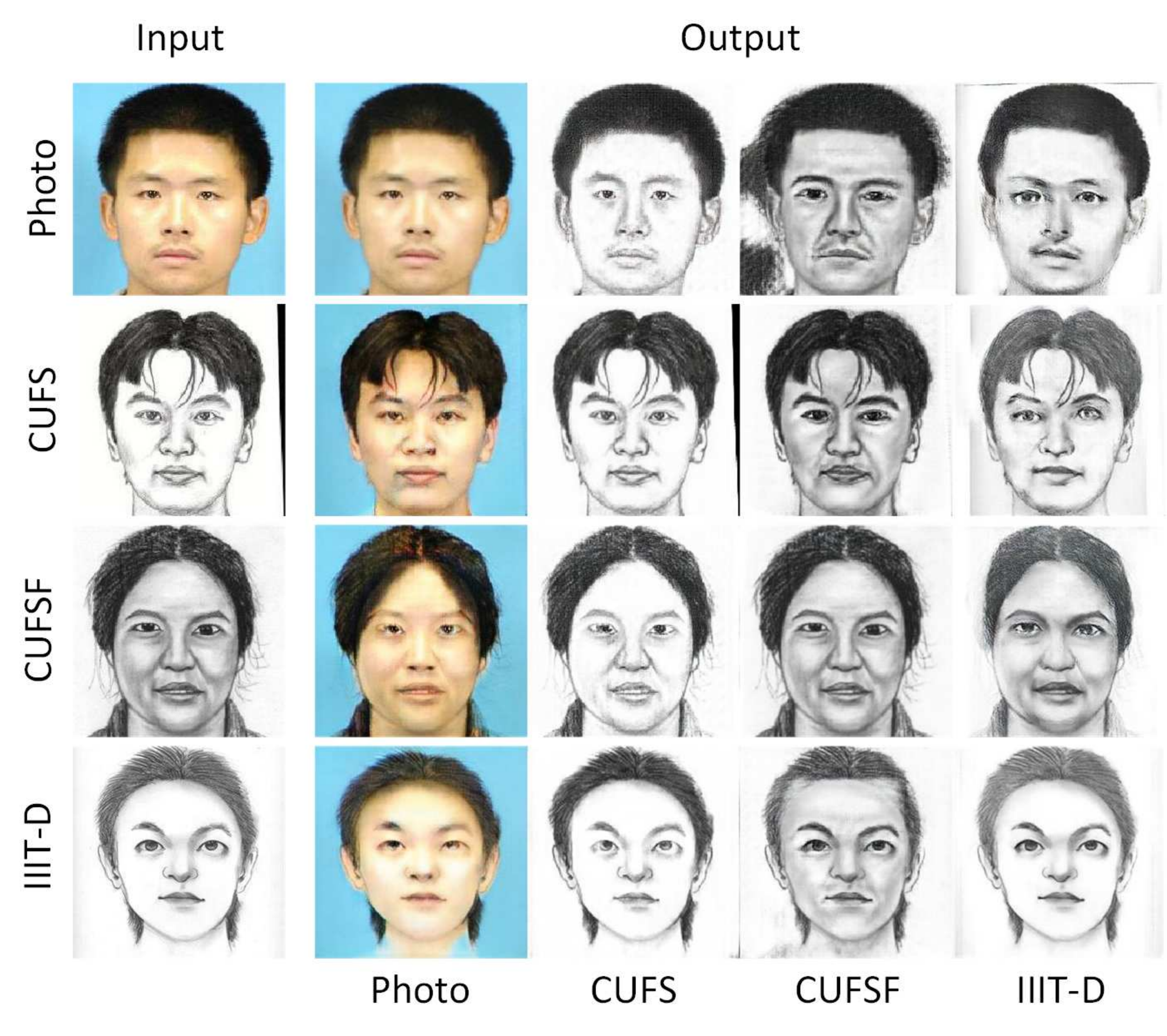}
	\caption{Example of face synthesis among face photos and three different styles of face sketches. Face photos are from the CUFS database, and face sketches are from CUFS ,CUFSF, IIIT-D respectively. }
	\label{figs:domain4}
\end{figure}

\begin{figure}
	\centering
	\includegraphics[width=0.9\hsize]{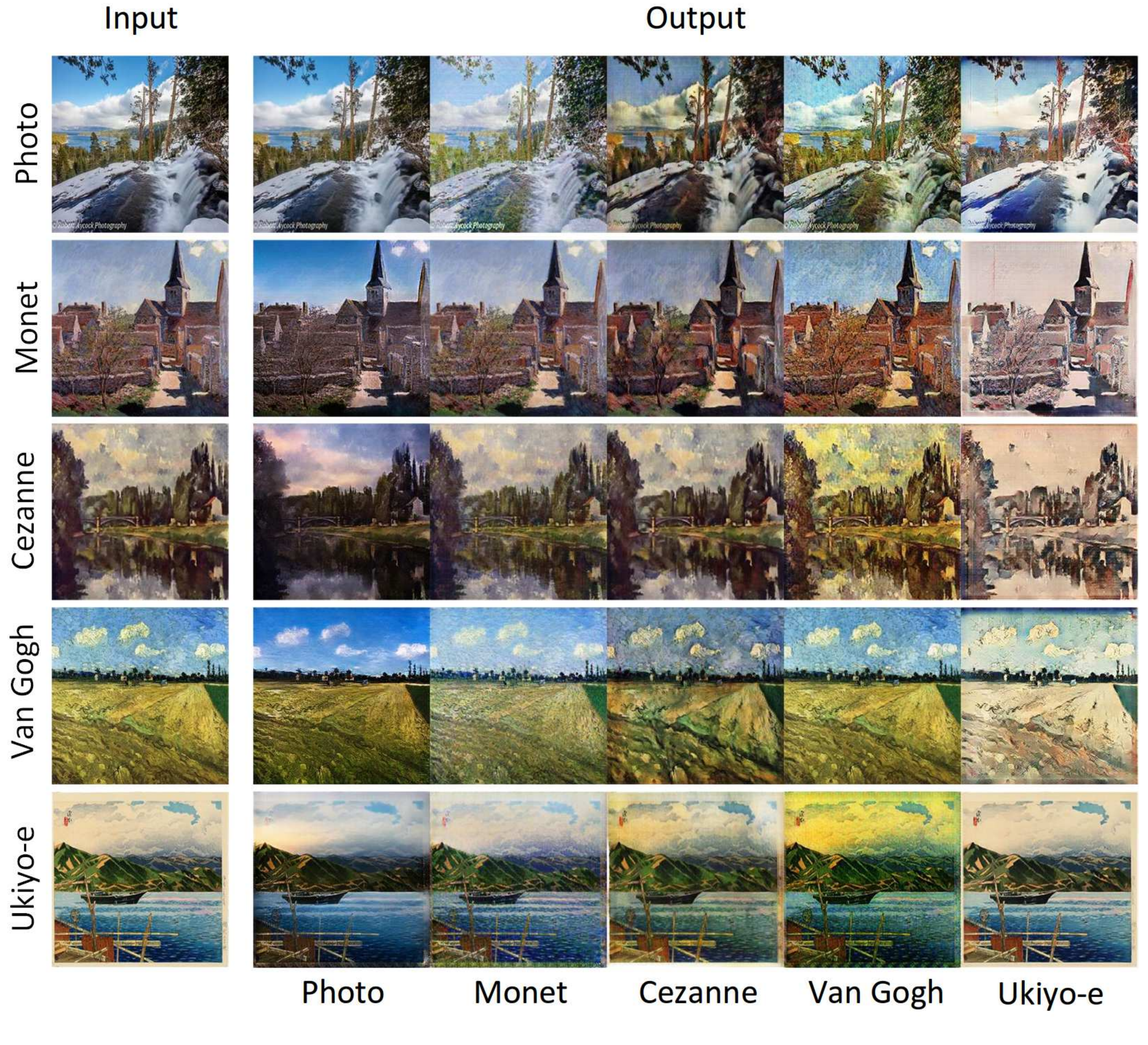}
	\caption{Example of style transfer among photographic, painting styles of Monet, Cezanne, Van Gogh, and Ukiyo-e. }
	\label{figs:domain5}
\end{figure}

\begin{figure*}
	\centering
	\includegraphics[width=0.9\hsize]{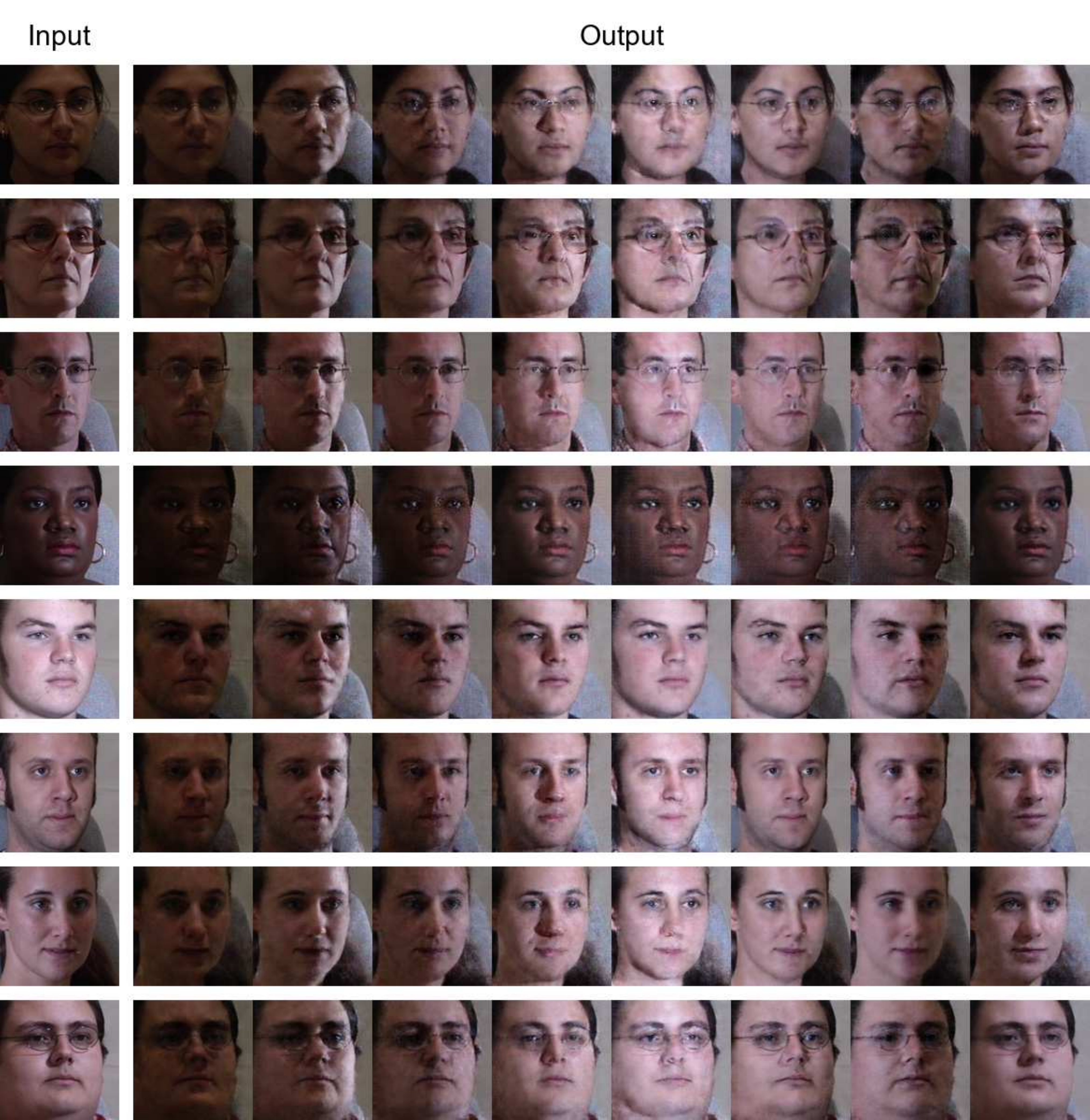}
	\caption{Example of face re-lighting among eight different illuminations, which successively change from dark, shadow, normal to another shadow. The left column is the inputs from different domains and the other right columns represent the outputs to different domains. The input row number corresponds to the output column number, and they have same domain label. }
	\label{figs:domain8}
\end{figure*}

We apply MDT to several tasks with different domain numbers, which includes semantic segmentation (\reffig{figs:domain2} (a)), object transfiguration (\reffig{figs:domain2} (b)), image colorization (\reffig{figs:domain3}), style transfer (\reffig{figs:domain4} and \reffig{figs:domain5}), and face synthesis (\reffig{figs:domain4} and \reffig{figs:domain8} ).

\reffig{figs:domain2} shows two bi-directional translations on CMP Facades \cite{Tylecek13} and a fruit collection of apples and oranges collected from ImageNet \cite{deng2009imagenet} by zhu et.al \cite{zhu2017unpaired}.

\reffig{figs:domain3} shows an example of translation among summer, winter and black-white (BW) photographs. The BW images are randomly selected from the two seasons and are converted to grayscale photos.

\reffig{figs:domain4} shows the results of the mutual translation among real faces and three different styles of face sketches. The face sketch databases are CUFS \cite{wang2009face}, CUFSF \cite{zhang2011coupled}, and the IIIT-D \cite{bhatt2012memetically}. We choose face photos from CUFS and combine them with face sketches in all databases to represent four different domains.

\reffig{figs:domain5} shows an example of another style transfer across five domains.

\reffig{figs:domain8} shows a part of results on another face re-lighting task which contains eight different face illuminations.

\end{document}